\def\year{2020}\relax
\documentclass[letterpaper]{article}
\usepackage{aaai20}
\usepackage{times}
\usepackage{helvet}
\usepackage{courier}
\usepackage{url}
\usepackage{graphicx}
\usepackage{booktabs}
\usepackage{subcaption}
\usepackage{amsmath}
\usepackage{amsfonts}
\usepackage{mathtools}
\usepackage{multirow}
\usepackage{enumitem}  
\usepackage{pifont}

\newcommand{\eg}{\emph{e.g.,}~}
\newcommand{\ie}{\emph{i.e.,}~}

\newcommand{\argmin}{\operatornamewithlimits{argmin}}

\title{Fine-Grained Fashion Similarity Learning by \\ Attribute-Specific Embedding Network}
\author{
Zhe Ma\textsuperscript{\rm 1}\thanks{Zhe Ma, Jianfeng Dong and Yao Zhang are the co-first authors. },
Jianfeng Dong\textsuperscript{\rm 2,\rm 3}\footnotemark[1]\thanks{Corresponding authors: Jianfeng Dong and Shouling Ji.},
Yao Zhang\textsuperscript{\rm 1}\footnotemark[1],\\
\Large \textbf{Zhongzi Long\textsuperscript{\rm 1},
Yuan He\textsuperscript{\rm 4},
Hui Xue\textsuperscript{\rm 4},
Shouling Ji\textsuperscript{\rm 1,\rm 3}\footnotemark[2]}\\
\textsuperscript{\rm 1}Zhejiang University,
\textsuperscript{\rm 2}Zhejiang Gongshang University,\\
\textsuperscript{\rm 3}Alibaba-Zhejiang University Joint Institute of Frontier Technologies,
\textsuperscript{\rm 4}Alibaba Group\\
\{maryeon, y.zhang, akasha, sji\}@zju.edu.cn,
dongjf24@gmail.com,
\{heyuan.hy, hui.xueh\}@alibaba-inc.com
}

\begin{document}
\maketitle
\begin{abstract}
This paper strives to learn fine-grained fashion similarity. In this similarity paradigm, one should pay more attention to the similarity in terms of a specific design/attribute among fashion items, which has potential values in many fashion related applications such as fashion copyright protection. To this end, we propose an \textit{Attribute-Specific Embedding Network (ASEN)} to jointly learn multiple attribute-specific embeddings in an end-to-end manner, thus measure the fine-grained similarity in the corresponding space. With two attention modules, \ie \textit{Attribute-aware Spatial Attention} and \textit{Attribute-aware Channel Attention}, ASEN is able to locate the related regions and capture the essential patterns under the guidance of the specified attribute, thus make the learned attribute-specific embeddings better reflect the fine-grained similarity. Extensive experiments on four fashion-related datasets show the effectiveness of ASEN for fine-grained fashion similarity learning and its potential for fashion reranking. 
Code and data are available at https://github.com/Maryeon/asen.
\end{abstract}

\section{Introduction} \label{sec:intro}

\begin{figure}[tb!]
\centering\includegraphics[width=0.9\columnwidth]{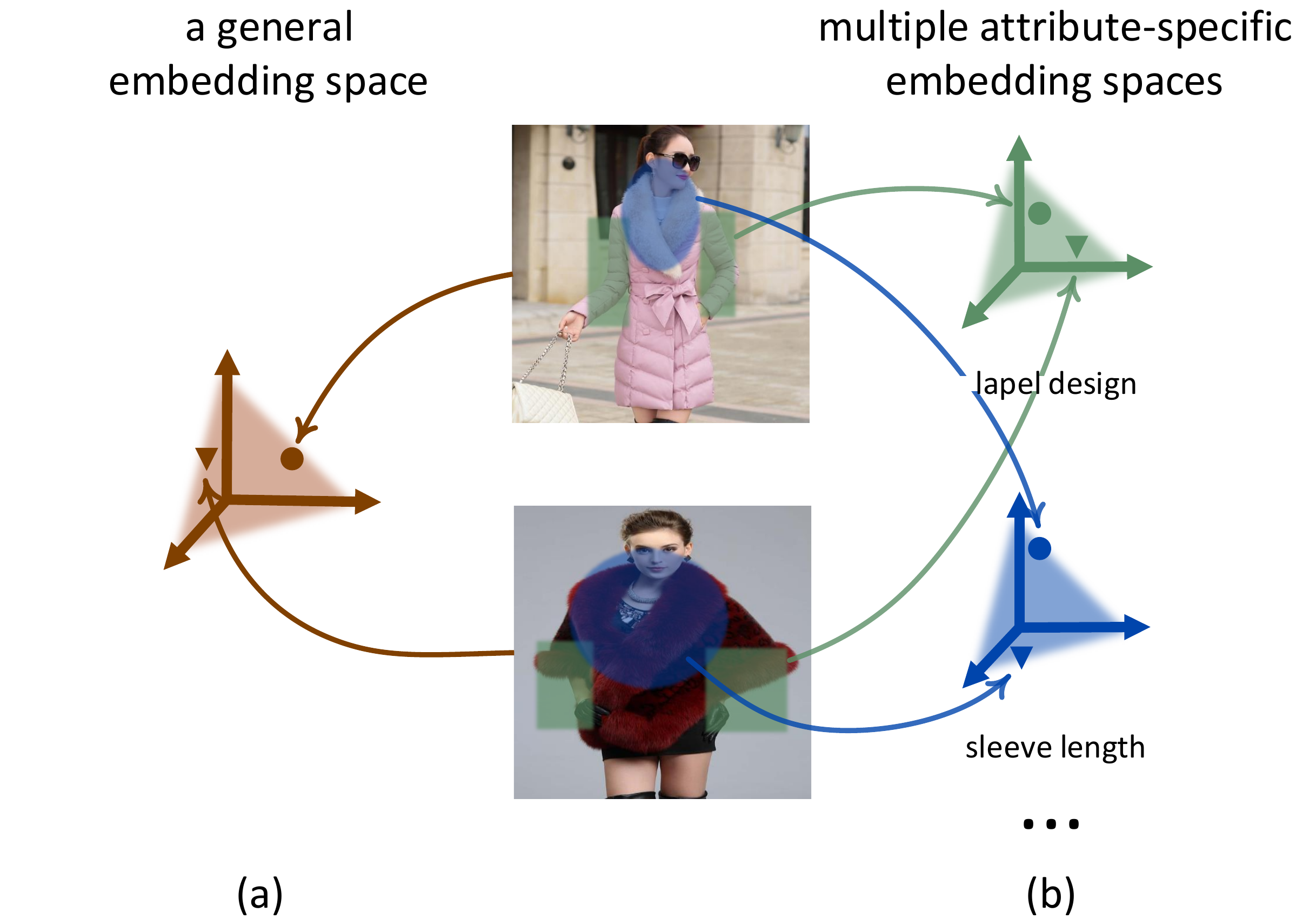}
\caption{As fashion items typically have various attributes, we propose to learn multiple attribute-specific embeddings, thus the fine-grained similarity can be better reflected in the corresponding attribute-specific space.}\label{fig:concept}
\end{figure}

Learning the similarity between fashion items is essential for a number of fashion-related tasks including in-shop clothes retrieval \cite{liu2016deepfashion,ak2018efficient}, cross-domain fashion retrieval \cite{huang2015DARNdataset,ji2017cross}, fashion compatibility prediction \cite{he2016learning,vasileva2018learning} and so on.
The majority of methods are proposed to learn a general embedding space so the similarity can be computed in the space \cite{zhao2017memory,ji2017cross,han2017learning}.
As the above tasks aim to search for identical or similar/compatible fashion items w.r.t. the query item, methods for these tasks tend to focus on the overall similarity.
In this paper, we aim for the fine-grained fashion similarity. Consider the two fashion images in Fig. \ref{fig:concept}, although they appear to be irrelevant overall, they actually present similar characteristics over some attributes, \eg both of them have the similar lapel design. We consider such similarity in terms of a specific attribute as the fine-grained similarity.

There are cases where one would like to search for fashion items with certain similar designs instead of identical or overall similar items, so the fine-grained similarity matters in such cases.
In the fashion copyright protection scenario \cite{martin2019fashion}, the fine-grained similarity is also important to find items with plagiarized designs. Hence, learning the fine-grained similarity is necessary.
However, to the best of our knowledge, such a similarity paradigm has been ignored by the community to some extent, only one work focuses on it. 
In \cite{veit2017}, they first learn an overall embedding space, and then employ a fixed mask to select relevant embedding dimensions w.r.t. the specified attribute. The fine-grained similarity is measured in terms of the masked embedding feature.
In this work, we go further in this direction. 
As shown in Fig. \ref{fig:concept}, we propose to learn multiple attribute-specific embedding spaces thus measure the fine-grained similarity in the corresponding space. 
For example, from the perspective of \textit{neckline design}, the similarity between two clothes can be measured in the embedding space of \textit{neckline design}.
To this end, we propose an \textit{Attribute-Specific Embedding Network (ASEN)} to jointly learn multiple attribute-specific embeddings in an end-to-end manner.
Specifically, we introduce the novel \textit{attribute-aware spatial attention (ASA)} and \textit{attribute-aware channel attention (ACA)} modules in the network, allowing the network being able to locate the related regions and capture the essential patterns w.r.t. the specified attribute.
It is worth pointing out that fine-grained similarity learning is orthogonal to overall similarity learning, allowing us to utilize ASEN to facilitate traditional fashion retrieval, such as in-shop clothes retrieval.
In sum, this paper makes the following contributions:
\begin{itemize}
\item Conceptually, we propose to learn multiple attribute-specific embedding spaces for fine-grained fashion similarity prediction. As such, a certain fine-grained similarity between fashion items can be measured in the corresponding space. 
\item We propose a novel ASEN model to effectively realize the above proposal. Combined with ACA and ASA, the network extracts essential features under the guidance of the specified attribute, which benefits the fine-grained similarity computation.
\item Experiments on FashionAI, DARN, DeepFashion and Zappos50k datasets demonstrate the effectiveness of proposed ASEN for fine-grained fashion similarity learning and its potential for fashion reranking.
\end{itemize}

\section{Related Work} \label{sec:rel-work}

\textbf{Fashion Similarity Learning}
To compute the similarity between fashion items, the majority of existing works \cite{liu2016deepfashion,Gajic2018CrossDomainFI,shankar2017deep,ji2017cross,huang2015DARNdataset} learn a general embedding space thus the similarity can be measured in the learned space by standard distance metric, \eg cosine distance.
For instance, in the context of in-shop clothes retrieval, \cite{liu2016deepfashion} employs a Convolutional Neural Network (CNN) to embed clothes into a single compact feature space.
Similarly, for the purpose of fashion compatibility prediction, \cite{veit2015learning} also utilize a CNN to map fashion items in an embedding space, thus predict whether two input fashion items are compatible in the space.
Different from the above methods that focus on the overall similarity (identical or overall similar/compatible), we study the fine-grained similarity in the paper.
\cite{veit2017} have made a first attempt in this direction. In their approach, an overall embedding space is first learned, and the fine-grained similarity is measured in this space with the fixed mask w.r.t. a specified attribute. 
By contrast, we jointly learn multiple attribute-specific embedding spaces, and measure the fine-grained similarity in the corresponding attribute-specific space.
It is worth noting that \cite{vasileva2018learning,he2016learning} also learn multiple embedding spaces, but they still focus on the overall similarity.

\textbf{Attention Mechanism}
Recently attention mechanism has become a popular technique and showed superior effectiveness in various research areas, such as computer vision \cite{woo2018cbam,wang2017residual,qiao2018exploring} and natural language processing \cite{vaswani2017attention,bahdanau2014neural}.
To some extent, attention can be regarded as a tool to bias the allocation of the input information.
As fashion images always present with complex backgrounds, pose variations, etc., attention mechanism is also common in the fashion domain \cite{ji2017cross,wang2017clothing,han2017automatic,ak2018learning,ak2018efficient}. 
For instance, \cite{ak2018efficient} use the prior knowledge of clothes structure to locate the specific parts of clothes.
However, their approach can be only used for upper-body clothes thus limits its generalization.
\cite{wang2017clothing} propose to learn a channel attention implemented by a fully convolutional network.
The above attentions are in a self-attention manner without explicit guidance for attention mechanism.
In this paper, we propose two attribute-aware attention modules, which utilize a specific attribute as the extra input in addition to a given image. The proposed attention modules capture the attribute-related patterns under the guidance of the specified attribute.
Note that \cite{ji2017cross} also utilize attributes to facilitate attention modeling,
but they use all attributes of fashion items and aim for learning a better discriminative fashion feature. 
By contrast, we employ each attribute individually to obtain more fine-grained attribute-aware feature for fine-grained similarity computation.

\section{Proposed Method} \label{sec:method}

\begin{figure*}[tb!]
\centering\includegraphics[width=0.9\textwidth]{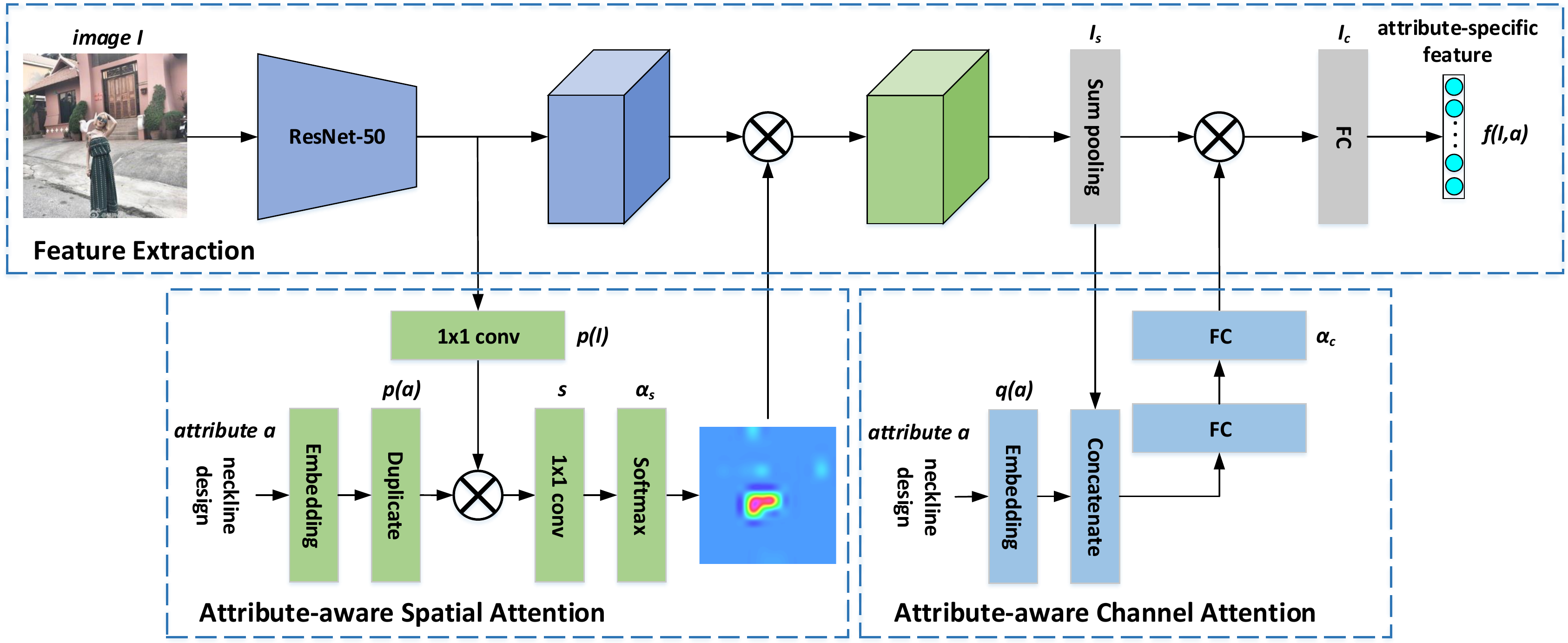}
\caption{The structure of our proposed Attribute-Specific Embedding Network (ASEN). Mathematical notations by the side of function blocks (\eg $\alpha_c$ on the right of FC layer) denotes their output.
}\label{fig:framework}
\end{figure*}

\subsection{Network Structure}
Given an image $I$ and a specific attribute $a$, we propose to learn an attribute-specific feature vector $f(I,a) \in \mathbb{R}^c$ which reflects the characteristics of the corresponding attribute in the image.
Therefore, for two fashion images $I$ and $I'$, the fine-grained fashion similarity w.r.t. the attribute $a$ can be expressed by the cosine similarity between $f(I,a)$ and $f(I',a)$.
Moreover, the fine-grained similarity for multiple attributes can be computed by summing up the similarity scores on the individual attributes. 
Note that the attribute-specific feature vector resides in the corresponding attribute-specific embedding space.
If there are $n$ attributes, $n$ attribute-specific embedding spaces can be learned jointly.
Fig. \ref{fig:framework} illustrates the structure of our proposed network.
The network is composed of a feature extraction branch combined with an attribute-aware spatial attention and an attribute-aware channel attention. For the ease of reference, we name the two attention modules as \textit{ASA} and \textit{ACA}, respectively.
In what follows, we first detail the input representation, followed by the description of two attribute-aware attention modules.

\subsubsection{Input Representation}
To represent the image, we employ a CNN model pre-trained on ImageNet \cite{deng2009imagenet} as a backbone network, \eg ResNet \cite{he2016deep}. 
To keep the spatial information of the image, we remove the last fully connected layers in the CNN. So the image is represented by $I \in \mathbb{R}^{c\times h\times w}$, where $h\times w$ is the size of the feature map, $c$ indicates the number of channels.
For the attribute, we represent it with a one-hot vector $a \in \{0,1\}^n$, where $n \in \mathbb{N}$ indicates the number of different attributes.

\subsubsection{Attribute-aware Spatial Attention (ASA)}
Considering the attribute-specific feature is typically related to the specific regions of the image, we only need to focus on the certain related regions. For instance, in order to extract the attribute-specific feature of the \textit{neckline design} attribute, the region around \textit{neck} is much more important than the others.
Besides, as fashion images always show up in large variations, \eg various poses and scales, using a fixed region with respect to a specific attribute for all images is not optimal.
Hence, we propose an attribute-aware spatial attention which adaptively attends to certain regions of the input image under the guidance of a specific attribute.
Given an image $I$ and a specific attribute $a$, we obtain the spatially attended vector w.r.t. the given attribute $a$ by $I_s = Att_{s}(I,a)$, where the attended vector is computed as the weighted average of input image feature vectors according to the given attribute.
Specifically, we first transform the image and the attribute to make their dimensionality same.
For the image, we employ a convolutional layer followed by a nonlinear $tanh$ activation function.
Formally, the mapped image $p(I)\in \mathbb{R}^{c'\times h\times w}$ is given by 
\begin{equation}
p(I) = tanh(Conv_{c'}(I)),
\end{equation}
where $Conv_{c'}$ indicates a convolutional layer that contains $c'$ $1\times 1$ convolution kernels.
For the attribute, we first project it into a $c'$-dimensional vector through an attribute embedding, implemented by a Fully Connected (FC) layer, then perform spatial duplication. Hence, the mapped attribute $p(a) \in \mathbb{R}^{c'\times h\times w}$ is
\begin{equation}
p(a) = tanh(W_a a)\cdot 1,
\end{equation}
Where $W_a\in \mathbb{R}^{c'\times n}$ denotes the transformation matrix and $1\in \mathbb{R}^{1\times h\times w}$ indicates spatially duplicate matrix.
After the feature mapping, the attention weights $\alpha_s \in \mathbb{R}^{h\times w} $ is computed as
\begin{equation}
\begin{array}{l}
s = tanh(Conv_{1}(p(a) \odot p(I))), \\
\alpha_s = softmax(s), \\
\label{eq:attention_map}
\end{array} 
\end{equation}
where $\odot$ indicates the element-wise multiplication, $Conv_{1}$ is a convolutional layer only containing one $1\times 1$ convolution kernel. Here, we employ a $softmax$ layer to normalize the attention weights.
With adaptive attention weights, the spatially attended feature vector of the image $I$ w.r.t. a specific attribute $a$ is calculated as:
\begin{equation}\label{eq:spatial_attention}
I_s = \sum_{j}^{h\times w} \alpha_{sj} I_j.
\end{equation}
where $\alpha_{sj} \in \mathbb{R}$ and $I_j \in \mathbb{R}^{c}$ are the attention weight and the feature vector at location $j$ of $\alpha_{s}$ and $I$, respectively.

\subsubsection{Attribute-aware Channel Attention (ACA)}
Although the attribute-aware spatial attention adaptively focuses on the specific regions in the image, the same regions may still be related to multiple attributes. For example, attributes \textit{collar design} and \textit{collar color} are all associated with the region around \textit{collar}.
Hence, we further employ attribute-aware channel attention over the spatially attended feature vector $I_s$.
The attribute-aware channel attention is designed as an element-wise gating function which selects the relevant dimensions of the spatially attended feature with respect to the given attribute.
Concretely, we first employ an attribute embedding layer to embed attribute $a$ into an embedding vector with the same dimensionality of $I_s$, that is:
\begin{equation}
q(a) = \delta(W_c a)
\end{equation}
where $W_c\in \mathbb{R}^{c\times n}$ denotes the embedding parameters and $\delta$ refers to \textit{ReLU} function. Note we use separated attribute embedding layers in \textit{ASA} and \textit{ACA}, considering the different purposes of the two attentions.
Then the attribute and the spatially attended feature are fused by simple concatenation, and further fed into the subsequent two FC layers to obtain the attribute-aware channel attention weights.
As suggested in \cite{hu2018squeeze}, we implement the two FC layers by a dimensionality-reduction layer with reduction rate $r$ and a dimensionality-increasing layer, which have fewer parameters than one FC layer. 
Formally, the attention weights $\alpha_c \in \mathbb{R}^c$ is calculated by:
\begin{equation}\label{eq:channel_attention}
    \alpha_c = \sigma(W_2 \delta(W_1 [q(a),I_s])),
\end{equation}
where $[,]$ denotes concatenation operation, $\sigma$ indicates \textit{sigmoid} function, $W_1\in \mathbb{R}^{\frac{c}{r}\times 2c}$ and $W_2\in \mathbb{R}^{c\times \frac{c}{r}}$ are transformation matrices. Here we omit the bias terms for description simplicity. 
The final output of the ACA is obtained by scaling $I_s$ with the attention weight $\alpha_c$:
\begin{equation}
    I_c = I_s \odot \alpha_c.
\end{equation}

Finally, we further employ a FC layer over $I_c$ to generate the attribute-specific feature of the given image $I$ with the specified attribute $a$:
\begin{equation}
f(I,a) = W I_c + b,
\end{equation}
where $W\in \mathbb{R}^{c\times c}$ is the transformation matrix, $b\in \mathbb{R}^c$ indicates the bias term.

\subsection{Model Learning}
We would like to achieve multiple attribute-specific embedding spaces where the distance in a particular space is small for images with the same specific attribute value, but large for those with the different ones.
Consider the \textit{neckline design} attribute for instance, we expect the fashion images with \textit{Round Neck} near those with the same \textit{Round Neck} in the \textit{neckline design} embedding space, but far away from those with \textit{V Neck}.
To this end, we choose to use the triplet ranking loss which is consistently found to be effective in multiple embedding learning tasks \cite{vasileva2018learning,dong2019dual}.
Concretely, we first construct a set of triplets $\mathcal{T} = \{(I, I^+, I^-| a)\}$, where $I^+$ and $I^-$ indicate images relevant and irrelevant with respect to image $I$ in terms of attribute $a$.
Given a triplet of $\{(I, I^+, I^-| a)\}$, triplet ranking loss is defined as
\begin{equation}\label{eq:loss}
\mathcal{L}(I, I^+, I^- | a) = max\{0,m - s(I,I^+|a)+s(I,I^-|a)\},
\end{equation}
where $m$ represents the margin, empirically set to be 0.2, $s(I,I'|a)$ denotes the fine-grained similarity w.r.t. the attribute $a$ which can be expressed by the cosine similarity between $f(I,a)$ and $f(I',a)$.
Finally, we train the model to minimize the triplet ranking loss on the triplet set $\mathcal{T}$, and the overall objective function of the model is as:
\begin{equation}\label{eq:obj}
\argmin_{\theta} \sum_{(I, I^+, I^- | a)\in \mathcal{T}}  \mathcal{L}(I, I^+, I^- | a),
\end{equation}
where $\theta$ denotes all trainable parameters of our proposed network.

\section{Evaluation} \label{sec:eval}

\subsection{Experimental  Setup}
To verify the viability of the proposed attribute-specific embedding network for fine-grained fashion similarity computation, we evaluate it on the following two tasks.
(1) Attribute-specific fashion retrieval: Given a fashion image and a specified attribute, its goal is to search for fashion images of the same attribute value with the given image.
(2) Triplet relation prediction: Given a triplet of $\{I, I', I''\}$ and a specified attribute, the task is asked to predict whether the relevance between $I$ and $I'$ is larger than that between $I$ and $I''$ in terms of the given attribute.

\begin{table*} [tb!]
\renewcommand{\arraystretch}{1.2}
\caption{Performance of attribute-specific fashion retrieval on FashionAI. Our proposed ASEN model consistently outperforms the other counterparts for all attribute types.}
\label{tab:fashionAI}
\centering 
\resizebox{2\columnwidth}{!}{
\begin{tabular}{l*{9}{c}}
\toprule
\multirow{2}{*}{\textbf{Method}} & \multicolumn{8}{c}{\textbf{MAP for each attribute}} & \multirow{2}{*}{\textbf{MAP}} \\
\cmidrule(l){2-9}
 & skirt length & sleeve length & coat length & pant length & collar design & lapel design & neckline design & neck design \\
\cmidrule(l){1-10}
Random baseline                & 17.20 & 12.50 & 13.35 & 17.45 & 22.36 & 21.63 & 11.09 & 21.19 & 15.79 \\
Triplet network       & 48.38 & 28.14 & 29.82 & 54.56 & 62.58 & 38.31 & 26.64 & 40.02 & 38.52 \\
CSN                            & 61.97 & 45.06 & 47.30 & 62.85 & 69.83 & 54.14 & 46.56 & 54.47 & 53.52 \\
ASEN  w/o ASA               & 62.65 & 49.98 & 49.02 & 63.48 & 69.10 & 61.65 & 50.88 & 57.10 & 56.35\\
ASEN  w/o ACA               & 58.12 & 43.30 & 42.30 & 60.03 & 65.98 & 49.95 & 46.86 & 52.06 & 50.87\\
ASEN                           & \textbf{64.44} & \textbf{54.63} & \textbf{51.27} & \textbf{63.53} & \textbf{70.79} & \textbf{65.36} & \textbf{59.50} & \textbf{58.67} & \textbf{61.02}\\
\bottomrule
\end{tabular}
 }
\end{table*}

\begin{table*} [tb!]
\renewcommand{\arraystretch}{1.2}
\centering
\caption{Performance of attribute-specific fashion retrieval on DARN. AESN with both ACA and ASA again performs best.} 
\label{tab:DARN}
\centering 
\resizebox{2\columnwidth}{!}{
\begin{tabular}{l*{10}{c}}
\toprule
\multirow{2}{*}{\textbf{Method}} & \multicolumn{9}{c}{\textbf{MAP for each attribute}} & \multirow{2}{*}{\textbf{MAP}} \\
\cmidrule(l){2-10}
 &clothes category&clothes button&clothes color&clothes length&clothes pattern&clothes shape & collar shape&sleeve length&sleeve shape \\
\cmidrule(l){1-11}
Random baseline                & 8.49 & 24.45 & 12.54 & 29.90 & 43.26 & 39.76 & 15.22 & 63.03 & 55.54 & 32.26 \\
Triplet network       & 23.59 & 38.07 & 16.83 & 39.77 & 49.56 & 47.00 & 23.43 & 68.49 & 56.48 & 40.14 \\
CSN                            & 34.10 & 44.32 & 47.38 & 53.68 & 54.09 & 56.32 & 31.82 & 78.05 & 58.76 & 50.86 \\
ASEN  w/o ASA               & 33.94 & 45.37 & 48.56 & 54.36 & 53.83 & 57.33 & 32.78 & 77.77 & 59.32 & 51.39 \\
ASEN  w/o ACA               & 30.39 & 42.37 & 49.14 & 50.18 & 53.63 & 48.84 & 26.03 & 75.28 & 57.99 & 48.02 \\
ASEN                           & \textbf{36.69} & \textbf{46.96} & \textbf{51.35} & \textbf{56.47} & \textbf{54.49} & \textbf{60.02} & \textbf{34.18} & \textbf{80.11} & \textbf{60.04} & \textbf{53.31}\\
\bottomrule
 \end{tabular}
}
 \end{table*}

\begin{figure*}[tb!]
\centering\includegraphics[width=0.9\textwidth]{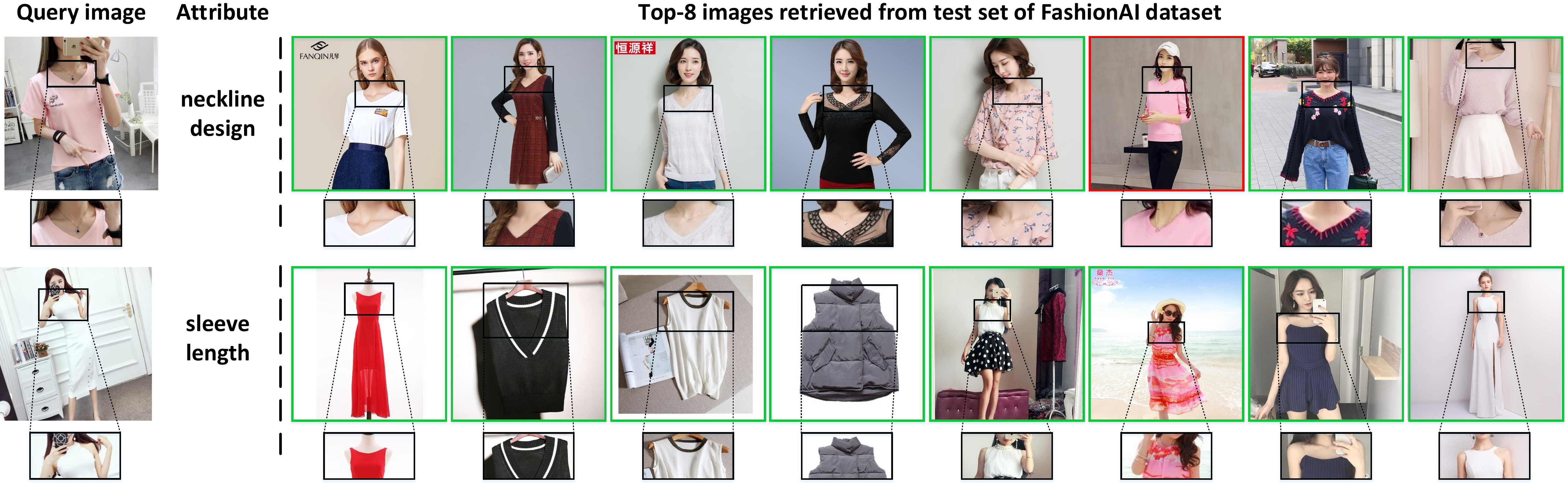}
\caption{Attribute-specific fashion retrieval examples on FashionAI.
Green bounding box indicates the image has the same attribute value with the given image in terms of the given attribute, while the red one indicates the different attribute values. The results demonstrate that our ASEN is good at capturing the fine-grained similarity among fashion items.
}\label{fig:retrieval_examples}
\end{figure*}

\subsubsection{Datasets}
As there are no existing datasets for attribute-specific fashion retrieval, we reconstruct three fashion datasets with attribute annotations to fit the task, \ie FashionAI \cite{zou2019fashionai}, DARN \cite{huang2015DARNdataset} and DeepFashion\cite{liu2016deepfashion}.
For triplet relation prediction, we utilize Zappos50k \cite{yu2014fine}.
\textit{FashionAI} is a large scale fashion dataset with hierarchical attribute annotations for fashion understanding. We choose to use the FashionAI dataset, because of its high-quality attribute annotations.
As the full FashionAI has not been publicly released, we utilize its early version released for the FashionAI Global Challenge 2018\footnote{https://tianchi.aliyun.com/markets/tianchi/FashionAI}.
The released FashionAI dataset consists of 180,335 apparel images, where each image is annotated with a fine-grained attribute. 
There are 8 attributes, and each attribute is associated with a list of attribute values. Take the attribute \textit{neckline design} for instance, there are 11 corresponding attribute values, such as \textit{round neckline} and \textit{v neckline}. 
We randomly split images into three sets by 8:1:1, which is 144k / 18k / 18k images for training / validation / test.
Besides, for every epoch, we construct 100k triplets from the training set for model training. Concretely, for a triplet with respect to a specific attribute, we randomly sample two images of the same corresponding attribute values as the relevant pair and an image with different attribute value as the irrelevant one. 
For validation or test set, 3600 images are randomly picked out as the query images, with remaining images annotated with the same attribute as the candidate images for retrieval. 
Additionally, we reconstruct DARN and DeepFashion in the same way as FashionAI. Details are included in the supplementary material.

\textit{Zappos50k} is a large shoe dataset consisting of 50,025 images collected from the online shoe and clothing retailer Zappos.com. For the ease of cross-paper comparison, we utilize the identical split provided by \cite{veit2017}.
Specifically, we use 70\% / 10\% / 20\% images for training / validation / test. 
Each image is associated with four attributes: the type of the shoes, the suggested gender of the shoes, the height of the shoes' heels and the closing mechanism of the shoes. For each attribute, 200k training, 20k validation and 40k testing triplets are sampled for model training and evaluation.

\subsubsection{Metrics}
For the task of attribute-specific fashion retrieval, we report the Mean Average Precision (MAP), a popular performance metric in many retrieval-related tasks \cite{awad2018trecvid,dong2018cross}. For the triplet relation prediction task, we utilize the prediction accuracy as the metric.

Due to the limited space of the paper, we present results on DeepFashion, implementation details and efficiency evaluation of our proposed model in the supplementary material.

\subsection{Attribute-Specific Fashion Retrieval}
Table \ref{tab:fashionAI} summarizes the performance of different models on FashionAI, and performance of each attribute type are also reported. As a sanity check, we also give the performance of a random baseline which sorts candidate images randomly. All the learning methods are noticeably better than the random result.
Among the five learning based models, the triplet network which learns a general embedding space performs the worst in terms of the overall performance, scoring the overall MAP of 38.52\%. The result shows that a general embedding space is suboptimal for fine-grained similarity computation.
Besides, our proposed ASEN outperforms CSN \cite{veit2017} with a clear margin.
We attribute the better performance to the fact that ASEN adaptively extracts feature w.r.t. the given attribute by two attention modules, while CSN uses a fixed mask to select relevant embedding dimensions.
Moreover, we investigate ASEN with a single attention, resulting in two reduced models, \ie ASEN w/o ASA and ASEN w/o ACA.
These two variants obtain the overall MAP of 56.35 and 50.87, respectively. The lower scores justify the necessity of both ASA and ACA attentions.
The result also suggests that attribute-aware channel attention is more beneficial.
Table \ref{tab:DARN} shows the results on the DARN dataset. Similarly, our proposed ASEN outperforms the other counterparts. The result again confirms the effectiveness of the proposed model for fine-grained fashion similarity computation.
Additionally, we also try the verification loss \cite{zheng2018discriminatively} in ASEN, but find its performance (MAP=50.63) worse than the triplet loss counterpart (MAP=61.02) on FashionAI. 
Some qualitative results of ASEN are shown in Fig. \ref{fig:retrieval_examples}. Note that the retrieved images appear to be irrelevant to the query image, as ASEN focuses on the fine-grained similarity instead of the overall similarity.
It can be observed that the majority of retrieved images share the same specified attribute with the query image. Consider the second example for instance, although the retrieved images are in various fashion category, such as \textit{dress} and \textit{vest}, all of them are \textit{sleeveless}. These results allow us to conclude that our model is able to figure out fine-grained patterns in images.

\subsection{Triplet Relation Prediction}

\begin{table} [tb!]
\renewcommand{\arraystretch}{1.2}
\caption{Performance of triplet relation prediction on Zappos50k. Our proposed ASEN is the best.}
\label{tab:zappos50k}
\centering 
\resizebox{0.8\columnwidth}{!}{
\begin{tabular}{lr}
\toprule
\textbf{Method} & \textbf{Prediction Accuracy(\%)}\\
\midrule
Random baseline           & 50.00 \\
Triplet network  & 76.28 \\
CSN                       & 89.27 \\
ASEN w/o ASA          & 90.18\\
ASEN w/o ACA          & 89.01 \\
ASEN                      & \textbf{90.79} \\
\bottomrule
\end{tabular}
 }
\end{table}

\begin{figure}[tb!]
\centering
\begin{subfigure}{0.28\columnwidth}
    \centering
    \includegraphics[width = \columnwidth]{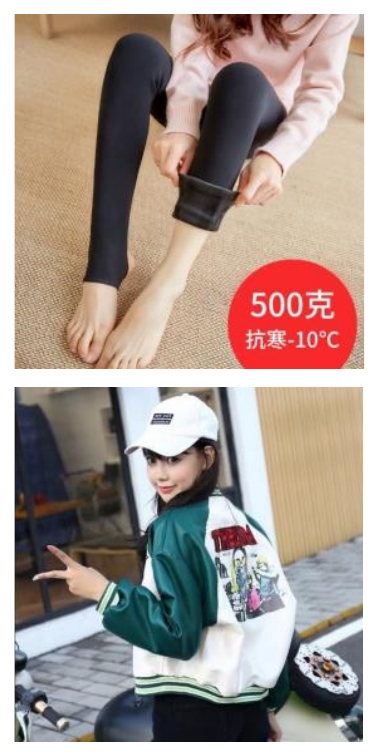}
    \caption{on FashionAI}
\end{subfigure}
\begin{subfigure}{0.28\columnwidth}
    \centering
    \includegraphics[width = \columnwidth]{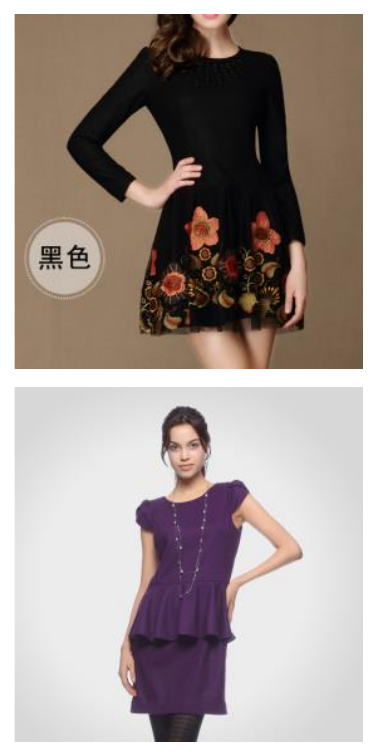}
    \caption{on DARN}
\end{subfigure}
\begin{subfigure}{0.28\columnwidth}
    \centering
    \includegraphics[width = \columnwidth]{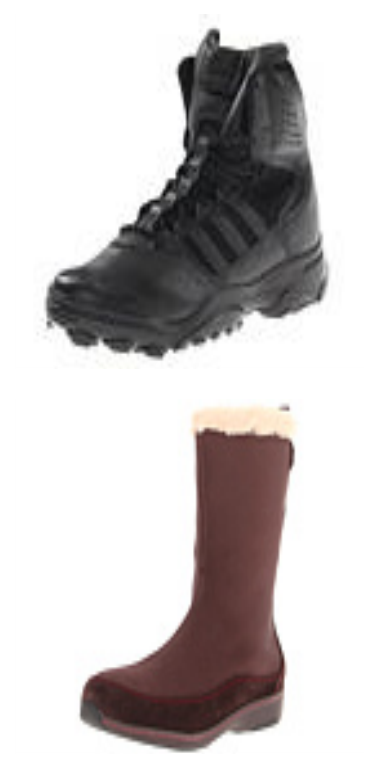}
    \caption{on Zappos50k}
\end{subfigure}
\caption{Images from FashionAI, DARN and Zappos50k, showing the images of Zappos50k are less challenging.}
\label{fig:examples}
\end{figure}

Table \ref{tab:zappos50k} shows the results on the Zappos50k dataset. 
Unsurprisingly, the random baseline achieves the worst performance as it predicts by random guess.
Among the four embedding learning models, our proposed model variants again outperform triplet network which only learns a general embedding space with a large margin. The result verifies the effectiveness of learning attribute-specific embeddings for triplet relation prediction.
Although ASEN is still better than its counterparts ASEN without ASA or ACA, its performance improvement is much less than that on FashionAI and DARN.
We attribute it to that images in Zappos50k are more iconic and thus easier to understand (see Fig. \ref{fig:examples}), so only ASA or ACA is enough to capture the fine-grained similarity for such ``easy" images.

\begin{figure}[tb!]
\centering
\begin{subfigure}{0.22\columnwidth}
    \centering
    \includegraphics[width = \columnwidth]{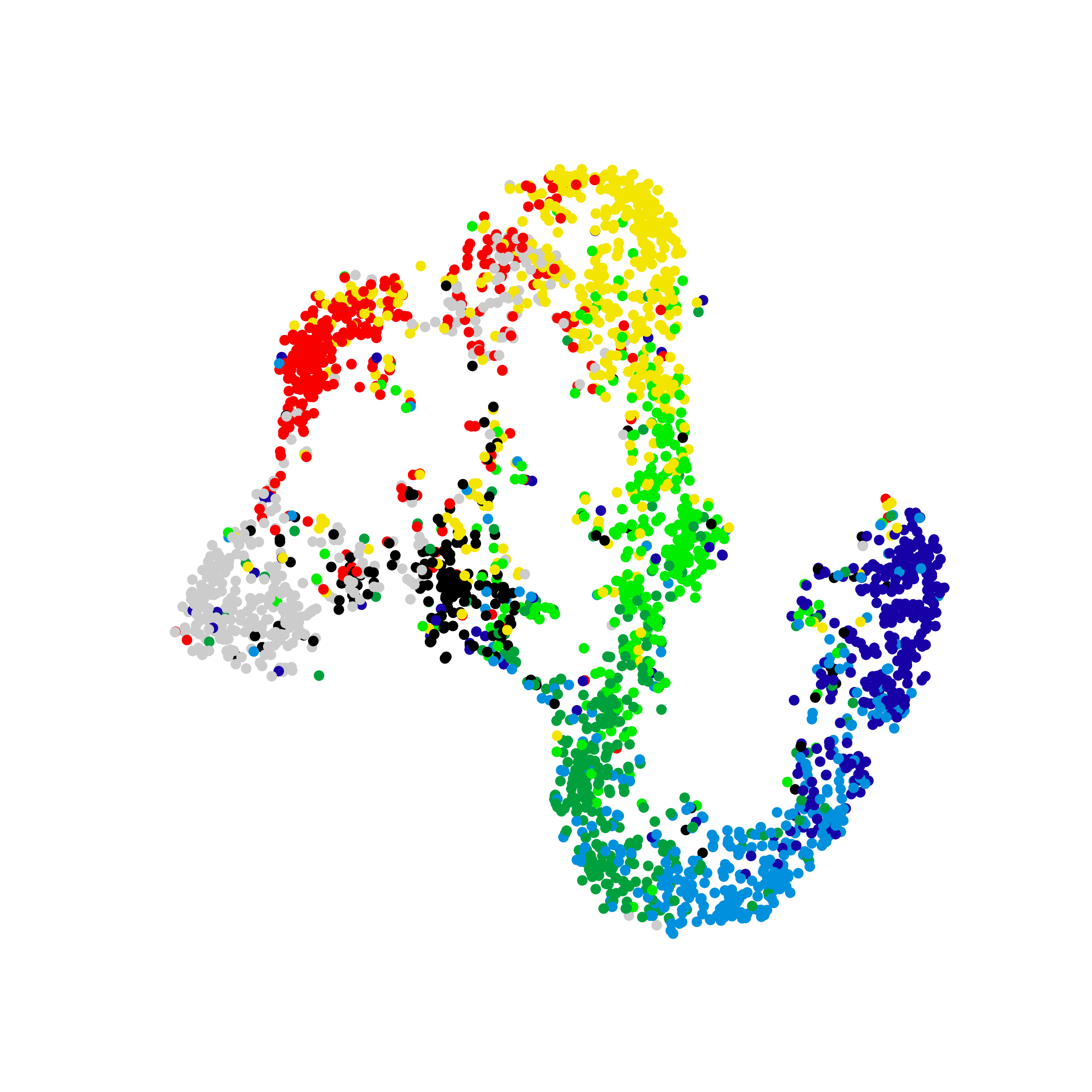}
    \caption{coat}
\end{subfigure}
\begin{subfigure}{0.22\columnwidth}
    \centering
    \includegraphics[width = \columnwidth]{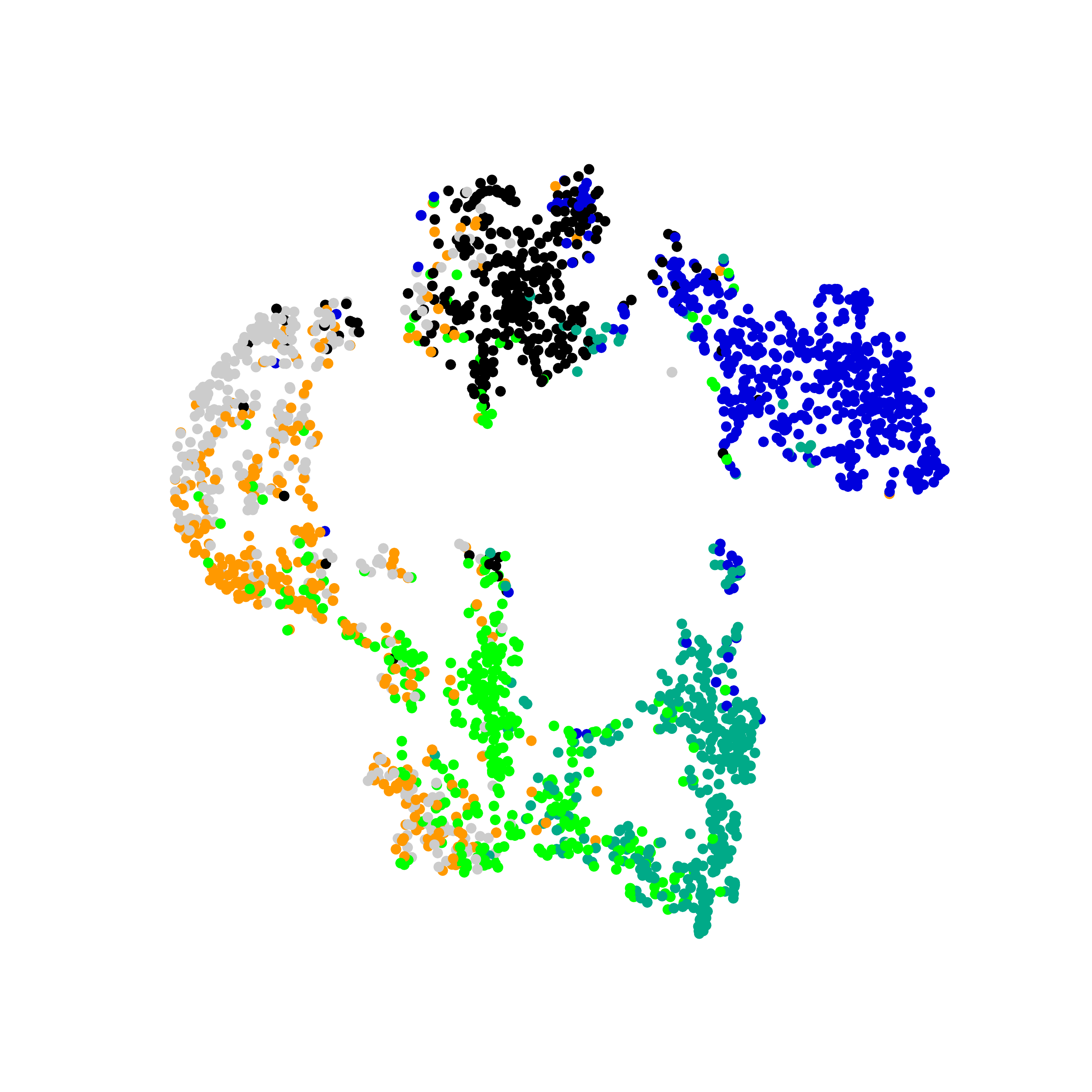}
    \caption{pant}
\end{subfigure}
\begin{subfigure}{0.22\columnwidth}
    \centering
    \includegraphics[width = \columnwidth]{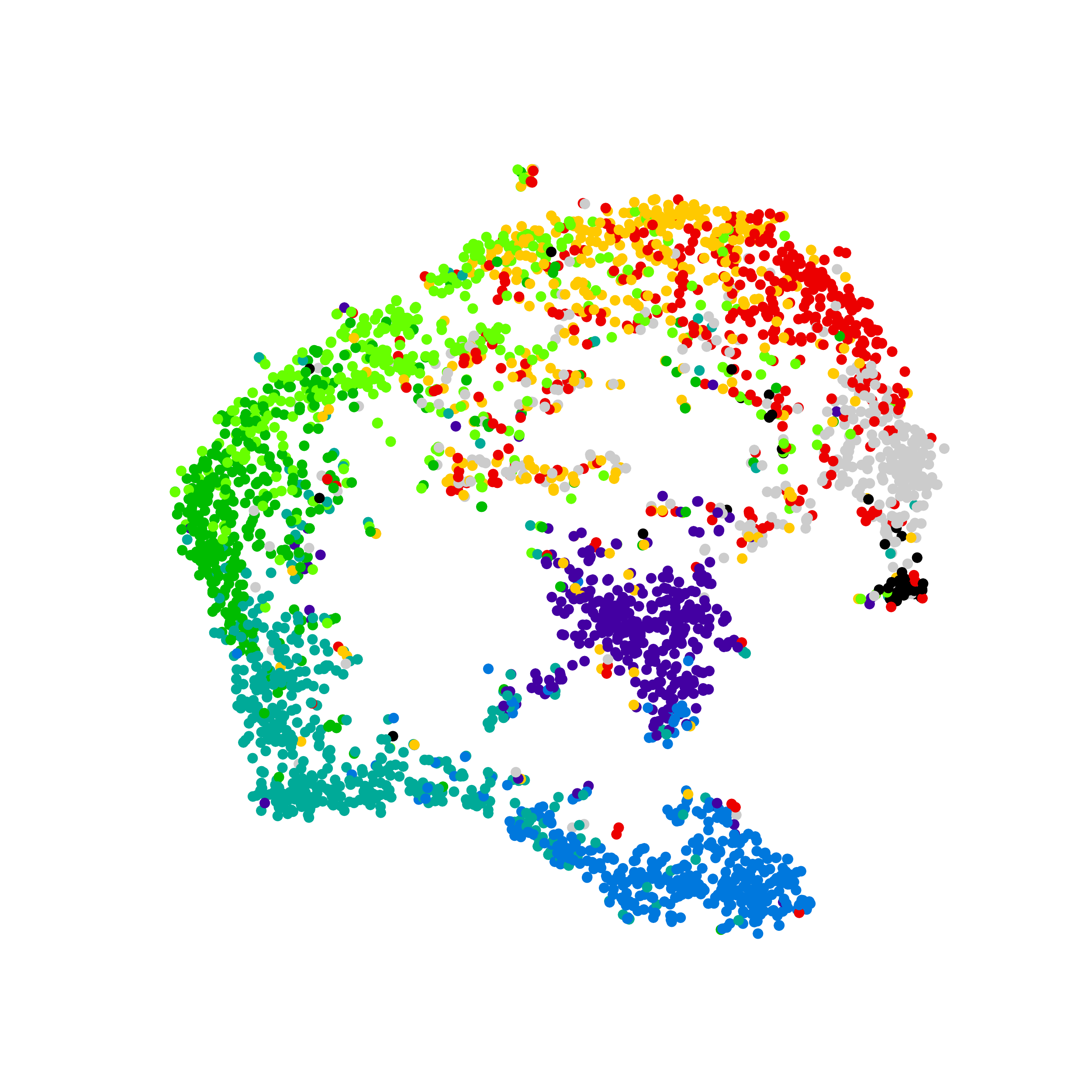}
    \caption{sleeve}
\end{subfigure}
\begin{subfigure}{0.22\columnwidth}
    \centering
    \includegraphics[width = \columnwidth]{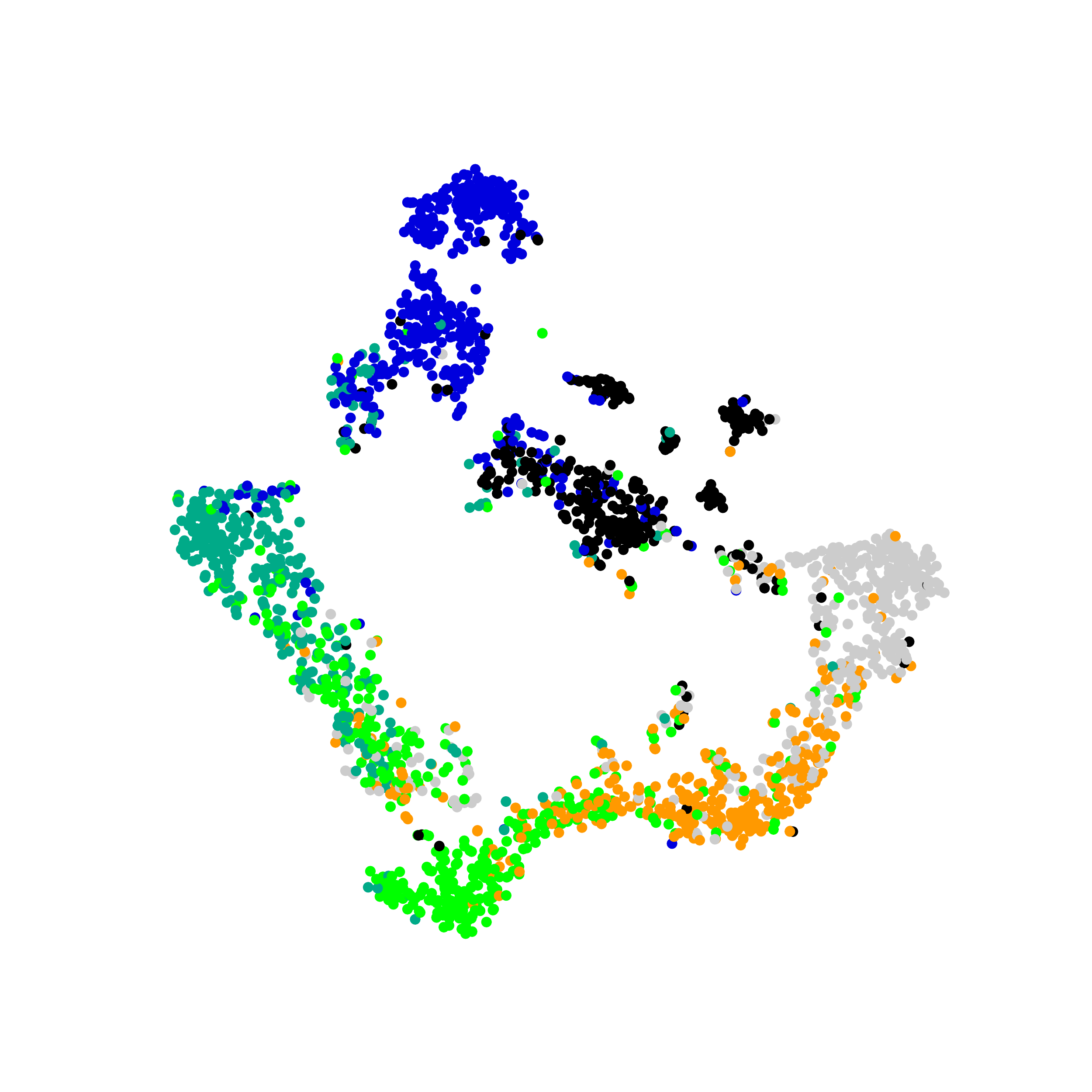}
    \caption{skirt}
\end{subfigure}
\begin{subfigure}{0.22\columnwidth}
    \centering
    \includegraphics[width = \columnwidth]{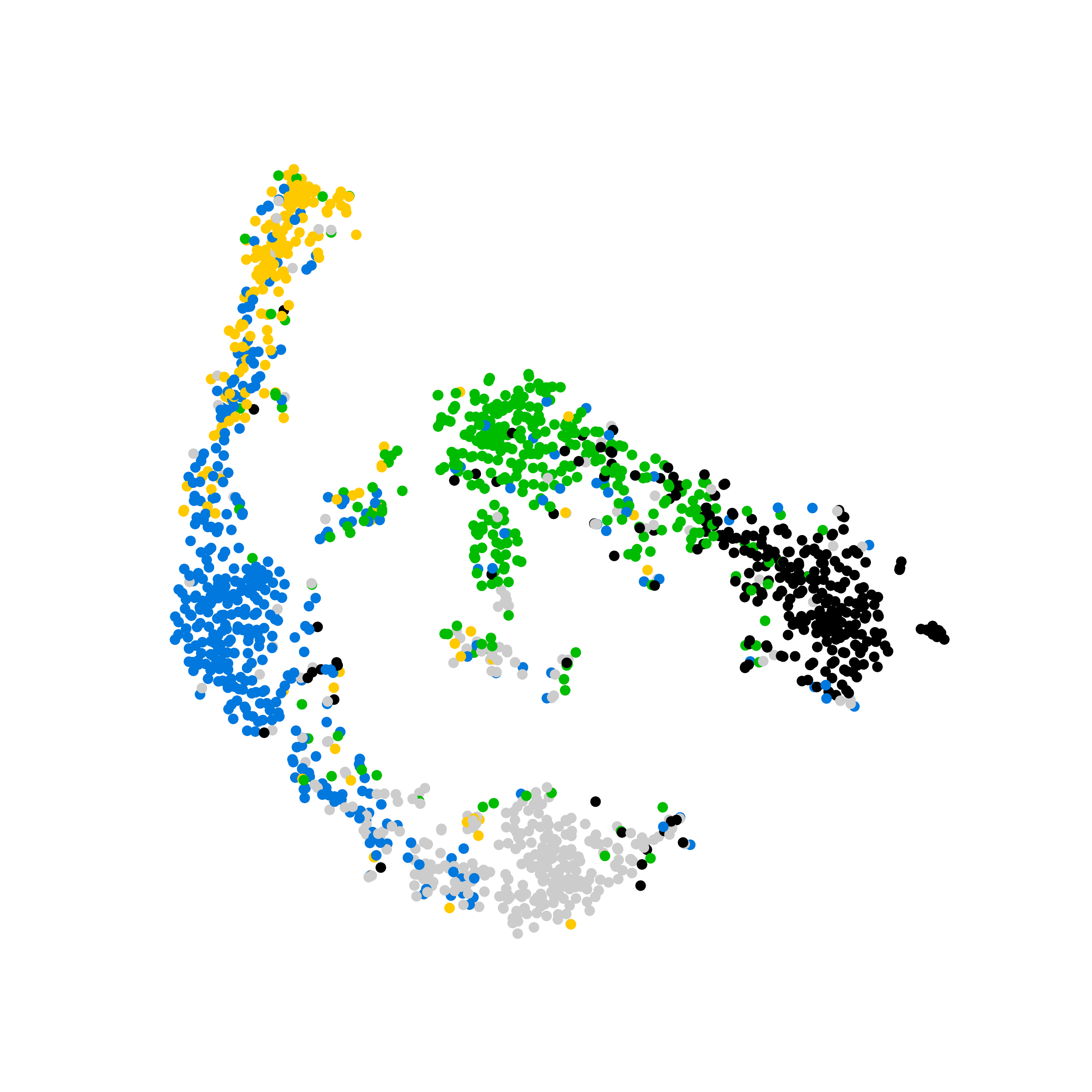}
    \caption{lapel}
\end{subfigure}
\begin{subfigure}{0.22\columnwidth}
    \centering
    \includegraphics[width = \columnwidth]{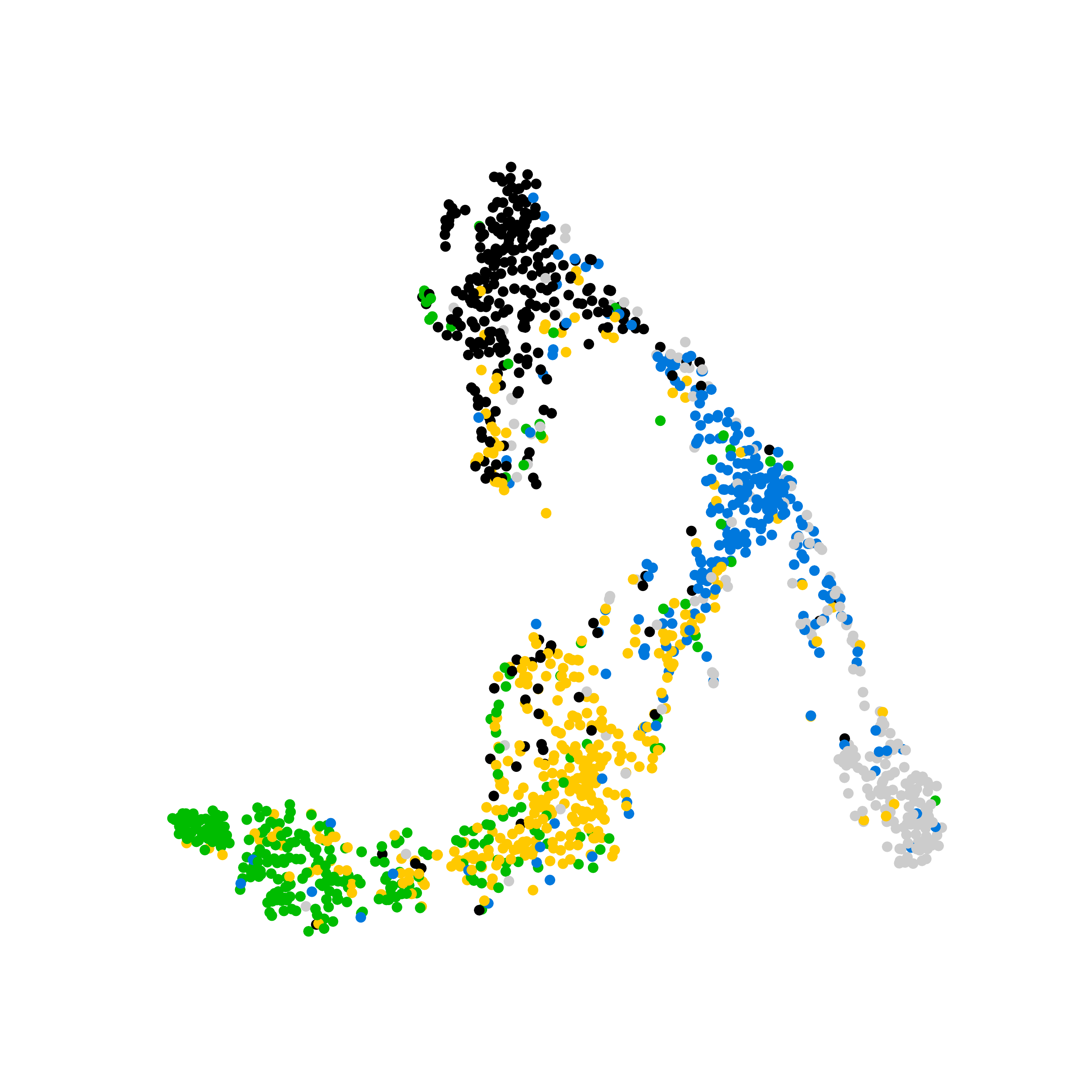}
    \caption{neck}
\end{subfigure}
\begin{subfigure}{0.22\columnwidth}
    \centering
    \includegraphics[width = \columnwidth]{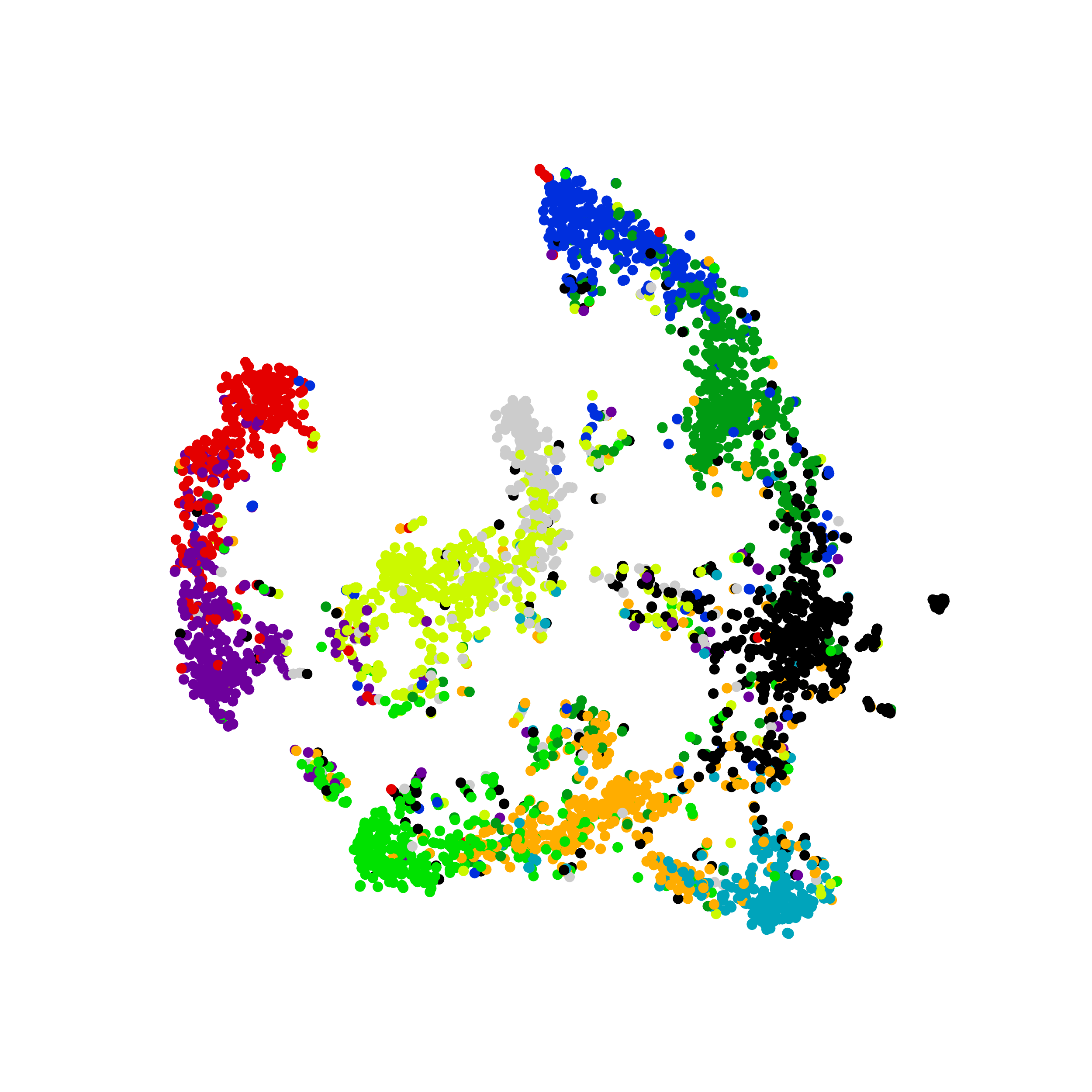}
    \caption{neckline}
\end{subfigure}
\begin{subfigure}{0.22\columnwidth}
    \centering
    \includegraphics[width = \columnwidth]{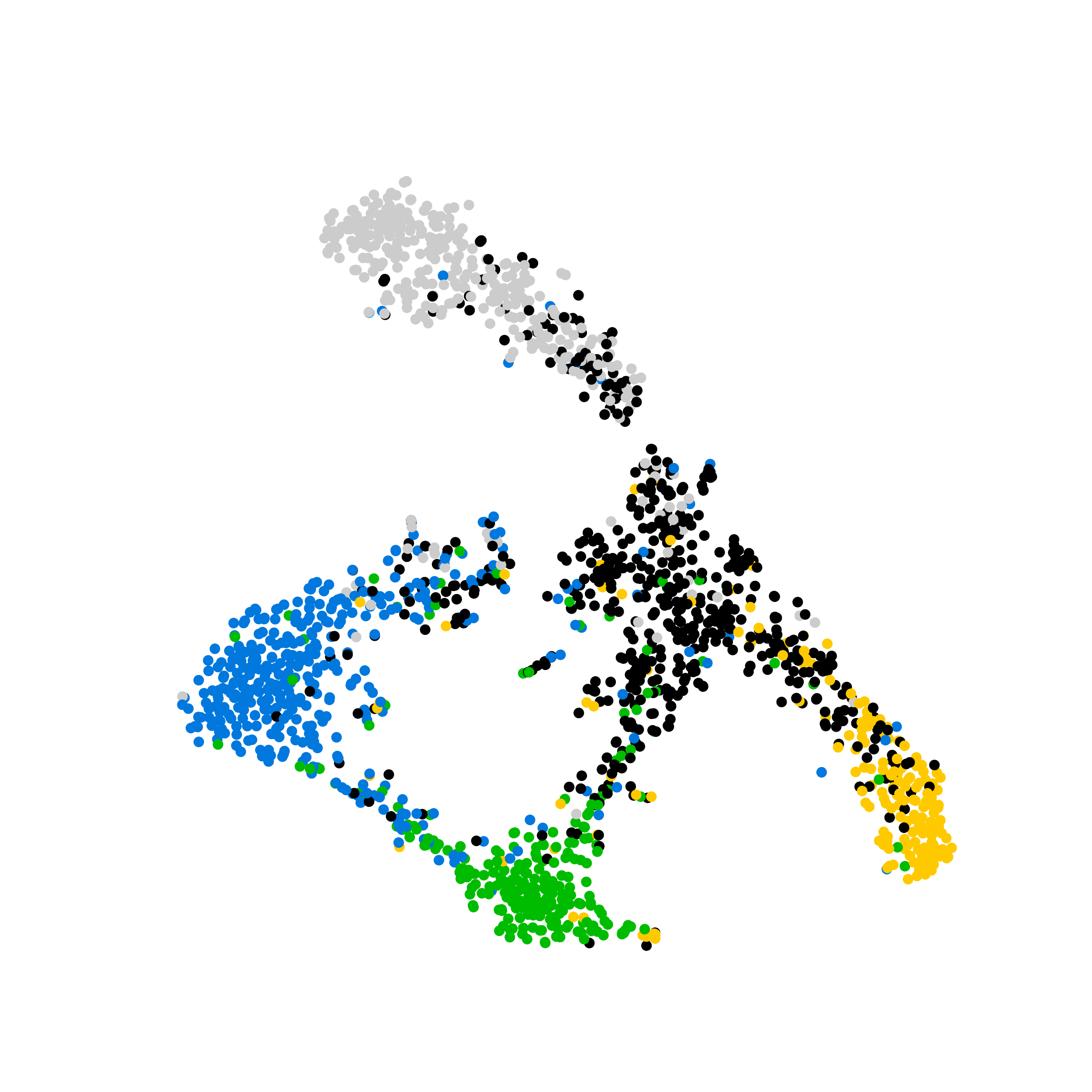}
    \caption{collar}
\end{subfigure}
\caption{t-SNE visualization of attribute-specific embedding spaces obtained by our proposed ASEN on FashionAI dataset. 
Dots with the same color indicate images annotated with the same attribute value. 
Best viewed in zoom in.}
\label{fig:tsne}
\end{figure}

\begin{figure}[tb!]
\centering
\includegraphics[width=0.45\columnwidth]{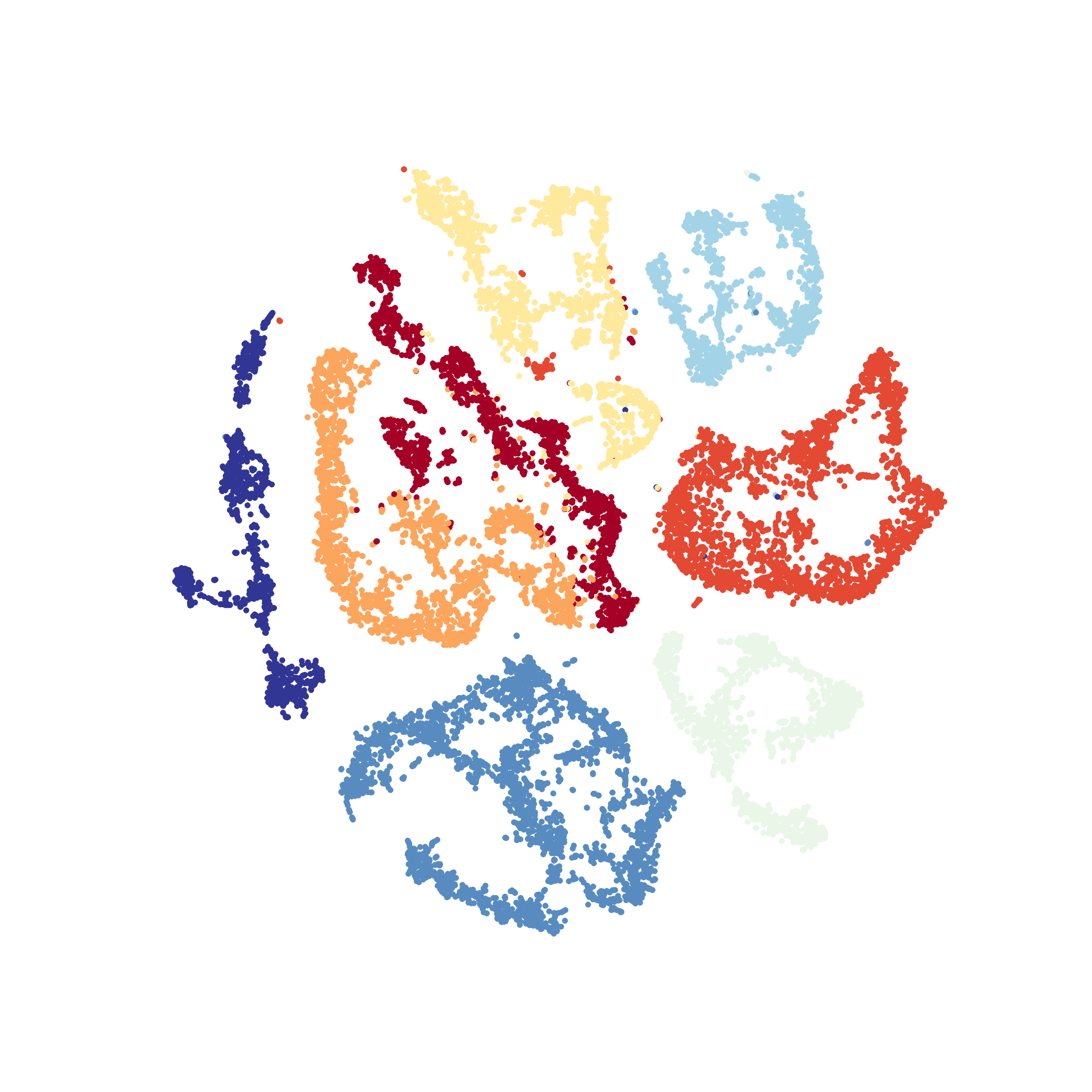}
\caption{t-SNE visualization of a whole embedding space comprised of eight attribute-specific embedding spaces learned by ASEN. Dots with the same color indicate images in the same attribute-specific embedding space. 
}\label{fig:tsne_attr_full_model}
\end{figure}

\subsection{What has ASEN Learned?}
\subsubsection{t-SNE Visualization}
\begin{figure*}[tb!]
\centering\includegraphics[width=0.88\textwidth]{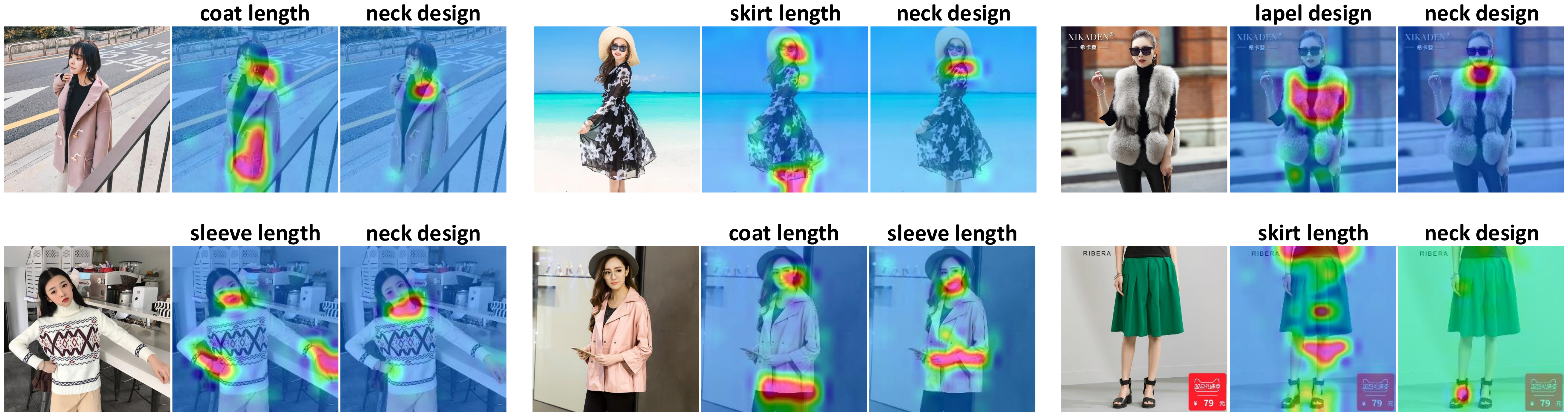}
\caption{Visualization of the attribute-aware spatial attention with the guidance of a specified attribute (above the attention image) on FashionAI.}\label{fig:spatial_attention}
\end{figure*}

\begin{figure*}[tb!]
\centering
\includegraphics[width=0.88\textwidth]{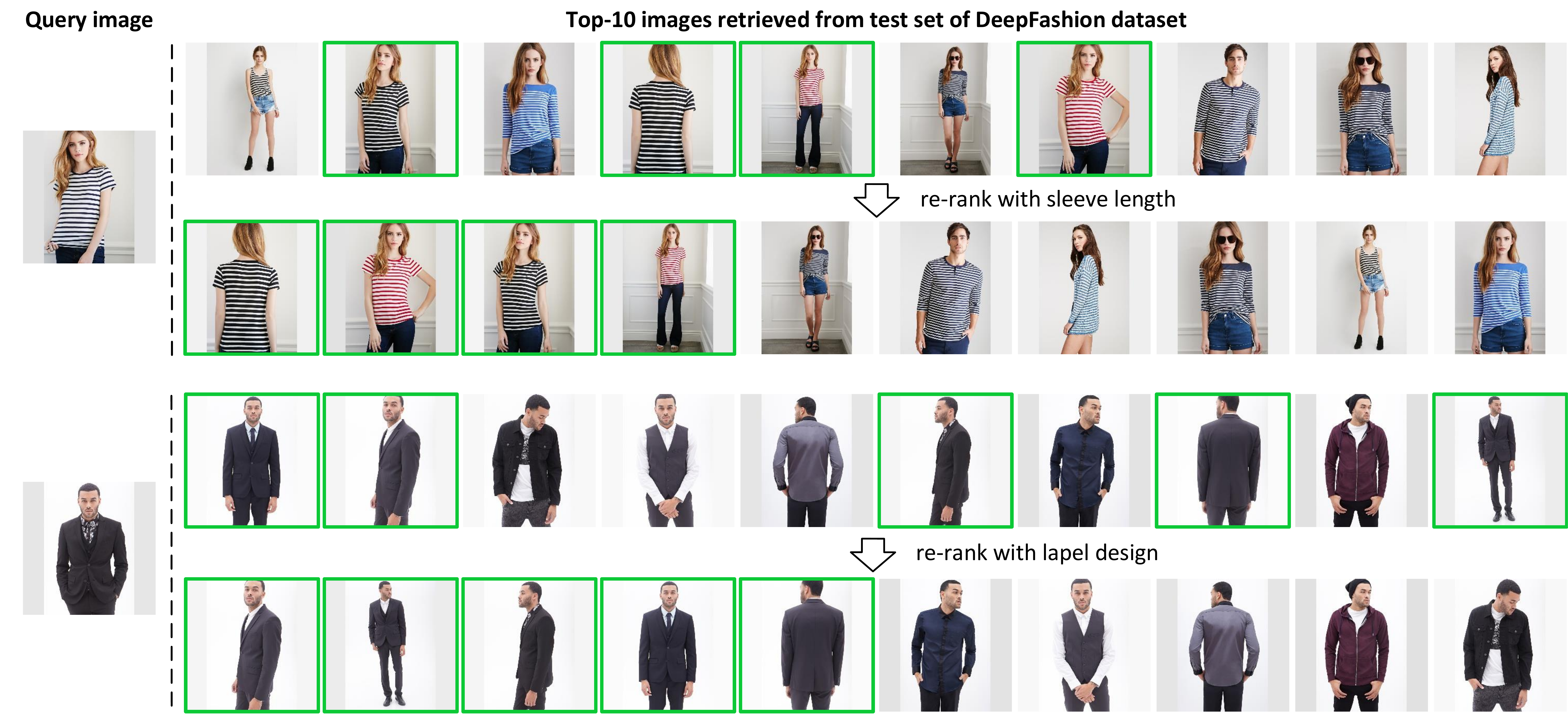}
\caption{Reranking examples for in-shop clothes retrieval on DeepFashion dataset. The ground-truth images are marked with green bounding box. After reranking by our proposed ASEN, the retrieval results become better. 
}\label{fig:resort}
\end{figure*}

In order to investigate what has the proposed ASEN learned, we first visualize the obtained attribute-specific embedding spaces. Specifically, we take all test images from FashionAI, and use t-SNE \cite{maaten2008visualizing} to visualize their distribution in 2-dimensional spaces.
Fig. \ref{fig:tsne} presents eight attribute-specific embedding spaces w.r.t. \textit{coat}, \textit{pant}, \textit{sleeve}, \textit{skirt length} and \textit{lapel}, \textit{neck}, \textit{neckline}, \textit{collar design} respectively. It is clear that dots with different colors are well separated and dots with the same color are more clustered in the particular embedding space. In other words, images with the same attribute value are close while images with different attribute value are far away. The result shows the good discriminatory ability of the learned attribute-specific embeddings by ASEN.
One may ask what is the relationship between the attribute-specific embedding spaces?
To answer this question, we visualize eight attribute-specific embeddings into a whole 2-dimensional space.
As shown in Fig. \ref{fig:tsne_attr_full_model}, different attribute-specific embeddings learned by ASEN are well separated.
The result is consistent with the fact that different attributes reflect different characteristics of fashion items.

\subsubsection{Attention Visualization}
To gain further insight of our proposed network, we visualize the learned attribute-aware spatial attention. As shown in Fig. \ref{fig:spatial_attention}, the learned attention map gives relative high responses on the relevant regions while low responses on irrelevant regions with the specified attribute, showing the attention is able to figure out which regions are more important for a specific attribute.
An interesting phenomenon can be observed that attention maps for length-related attributes are more complicated than that for design-related attributes; multiple regions show high response for the former. We attribute it to that the model requires to locate the start and end of a fashion item thus speculate its length.
Besides, consider the last example with respect to \textit{neck design} attribute, when the specified attribute \textit{neck design} can not be reflected in the image the attention response is almost uniform, which further demonstrates the effectiveness of attribute-aware spatial attention.

\subsection{The Potential for Fashion Reranking}
In this experiment, we explore the potential of ASEN for fashion reranking.
Specifically, we consider in-shop clothes retrieval task, in which given a query of in-shop clothes, the task is asked to retrieve the same items.
Triplet network is used as the baseline to obtain the initial retrieval result. The initial top 10 images are reranked in descending order by the fine-grained fashion similarity obtained by ASEN.
We train the triplet network on the official training set of DeepFashion, and directly use ASEN previously trained on FashionAI for the attribute-specific fashion retrieval task. Fig. \ref{fig:resort} presents two reranking examples. 
For the first example, by reranking in terms of the fine-grained similarity of \textit{sleeve length}, images have the same short sleeve with the query image are ranked higher, while the others with mid or long sleeve are ranked later.
Obviously, after reranking, the retrieval results become better. The result shows the potential of our proposed ASEN for fashion reranking.

\section{Summary and Conclusions} \label{sec:conc}

This paper targets at the fine-grained similarity in the fashion scenario.
We contribute an \textit{Attribute-Specific Embedding Network (ASEN)} with two attention modules, \ie ASA and ACA. ASEN jointly learns multiple attribute-specific embeddings, thus measuring the fine-grained similarity in the corresponding space. ASEN is conceptually simple, practically effective and end-to-end.
Extensive experiments on various datasets support the following conclusions. 
For fine-grained similarity computation, learning multiple attribute-specific embedding spaces is better than learning a single general embedding space.
ASEN with only ACA is more beneficial when compared with its counterpart with only ASA.
For state-of-the-art performance, we recommend ASEN with both attention modules.

\section{Acknowledgments}
This work was partly supported by the National Key Research and Development Program of China under No. 2018YFB0804102, NSFC under No. 61772466, 61902347 and U1836202, the Zhejiang Provincial Natural Science Foundation for Distinguished Young Scholars under No. LR19F020003, the Provincial Key Research and Development Program of Zhejiang, China under No. 2017C01055,
the Zhejiang Provincial Natural Science Foundation under No. LQ19F020002, and the Alibaba-ZJU Joint Research Institute of Frontier Technologies.

\bibliographystyle{aaai}
\bibliography{aaai20}

\appendix
\clearpage
\iffalse
\begin{figure*}[tb!]
\centering\includegraphics[width=2\columnwidth]{attr_tree}
\caption{\textbf{A hierarchical attributes tree as a structured understanding target}. Attributes that we said are the penultimate level of the tree, \eg collar, lapel, length of skirt \etal, which are highly relevant to certain regions on clothes.}\label{fig:attr_tree}
\end{figure*}

\begin{figure*}[tb!]
\centering\includegraphics[width=\columnwidth]{fashionAI_results}
\caption{\textbf{MAP evaluation on fashionAI dataset.}}\label{fig:fashionAI_results}
\end{figure*}
\fi

%\section{Evaluation}
\section{Experimental Details}

%=====================================================================================
\subsection{Dataset}
%=====================================================================================

\textit{DARN}\cite{huang2015DARNdataset} is constructed for attribute prediction and street-to-shop retrieval task. 
The dataset contains 253,983 images, each of which is annotated with 9 attributes. 
As some images' URLs have been broken, only 214,619 images are obtained for our experiments.
Images are randomly divided by 8:1:1 for training, validation and test, resulting in 171k,43k,43k images respectively.
Triplets are similarly sampled as on FashionAI dataset where two images with the same attribute values for an attribute form a positive pair and an image with different attribute value is randomly sampled as a negative one. 
We randomly choose 100k triplets for training.
For validation set and test set, images are split as query and candidate images by 1:4.

\textit{DeepFashion}\cite{liu2016deepfashion} is a large dataset which consists of four benchmarks for various tasks in the field of clothing including category and attribute prediction, in-shop clothes retrieval, fashion landmark detection and consumer-to-shop clothes retrieval. 
In our experiments, we use the category and attribute prediction benchmark for attribute-specific retrieval task and in-shop clothes retrieval benchmark for fashion reranking task.

The category and attribute prediction benchmark contains 289,222 images, 6 attributes and 1050 attribute values, and each image is annotated with several attributes.
Similar to DARN, we randomly split the images into training, validation and test set by 8:1:1 and construct 100k triplets for training. For validation set and test set, images are also split as query and candidate images by 1:4.

The in-shop clothes retrieval benchmark has a total of 52,712 images, consisting of 7,982 different items. Each item corresponds to multiple images by changing the color or angle of shot.
The images are officially divided into training, query, and gallery set, with 25k, 14k and 12k images.
We keep the training set unchanged in the experiment, and respectively divide the gallery and the query images to validation and test by 1:1.

Table \ref{tab:dataset} summarizes the attribute annotations and corresponding attribute values on the FashionAI, Zappos50k, DARN and category and attribute prediction benchmark of DeepFashion dataset.

\begin{figure}[tb!]
\centering
\begin{subfigure}{\columnwidth}
    \centering
    \includegraphics[width = \columnwidth]{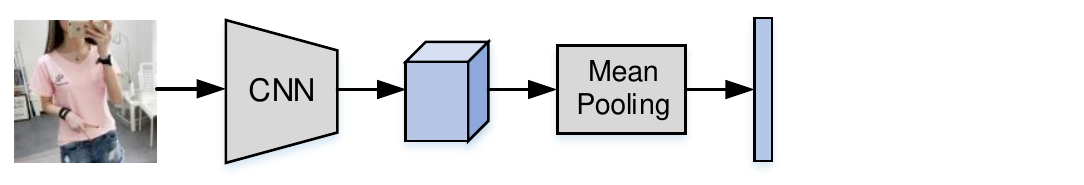}
    \caption{Triplet network}
    \label{fig:stdtriplet}
\end{subfigure}
\begin{subfigure}{\columnwidth}
    \centering
    \includegraphics[width = \columnwidth]{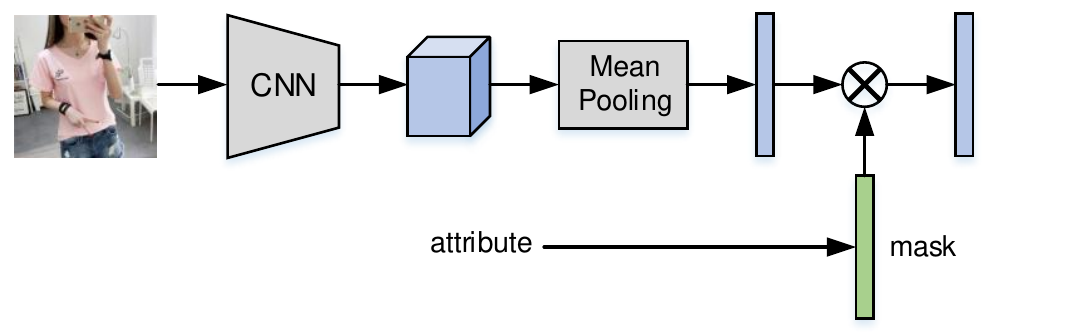}
    \caption{CSN}
    \label{fig:csn}
\end{subfigure}
\begin{subfigure}{\columnwidth}
    \centering
    \includegraphics[width = \columnwidth]{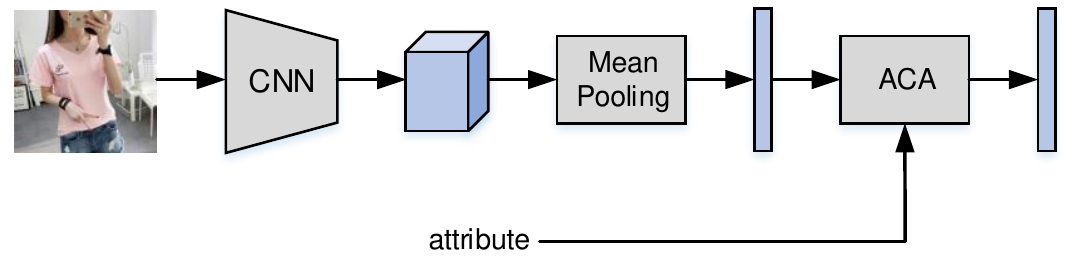}
    \caption{ASEN w/o ASA}
    \label{fig:asen_wo_asa}
\end{subfigure}
\begin{subfigure}{\columnwidth}
    \centering
    \includegraphics[width = \columnwidth]{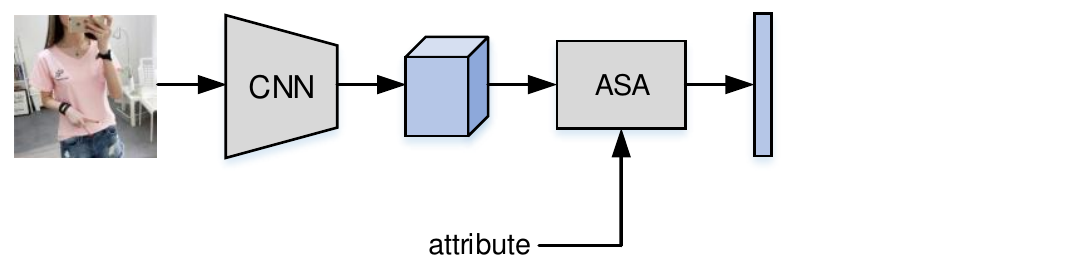}
    \caption{ASEN w/o ACA}
    \label{fig:asen_wo_aca}
\end{subfigure}
\begin{subfigure}{\columnwidth}
    \centering
    \includegraphics[width = \columnwidth]{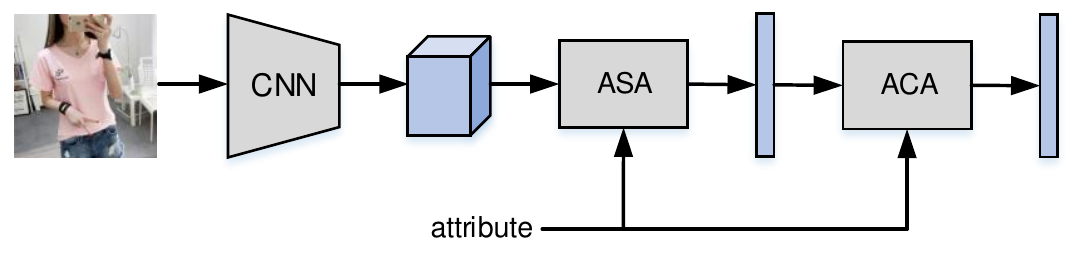}
    \caption{ASEN}
    \label{fig:asen}
\end{subfigure}
\caption{The conceptual structure of compared models and our proposed ASEN.}
\label{fig:models}
\end{figure}

\begin{table*} [tb!]
\renewcommand{\arraystretch}{1.2}
\centering
\caption{Performance of attribute-specific fashion retrieval on DeepFashion. Our proposed ASEN model consistently outperforms the other counterparts for all attribute types.}
% Existing works marked with stars (*) are our implementation.}
\label{tab:DeepFashion}
\centering 
\scalebox{0.9}{
\begin{tabular}{l*{6}{c}}
\toprule
\multirow{2}{*}{\textbf{Method}} & \multicolumn{5}{c}{\textbf{MAP for each attribute type}} & \multirow{2}{*}{\textbf{MAP}} \\
\cmidrule(l){2-6}
 &texture-related&fabric-related&shape-related&part-related&style-related \\
\cmidrule(l){1-7}
Random baseline                & 6.69 & 2.69 & 3.23 & 2.55 & 1.97 & 3.38 \\
Triplet network                & 13.26 & 6.28 & 9.49 & 4.43 & 3.33 & 7.36 \\
CSN                            & 14.09 & 6.39 & 11.07 & 5.13 & 3.49 & 8.01 \\
ASEN w/o ASA                   & 14.73 & 6.90 & 11.73 & 5.25 & 3.43 & 8.41 \\
ASEN w/o ACA                   & 14.71 & 6.78 & 11.71 & 5.03 & 3.42 & 8.32 \\
ASEN                           & \textbf{15.13} & \textbf{7.11} & \textbf{12.39} & \textbf{5.51} & \textbf{3.56} & \textbf{8.74}\\
\bottomrule
\end{tabular}
 }
\end{table*}

\linespread{1.2}
\begin{table*} [tb!]
%\linespread{1.5}
\centering
\caption{Summary of attribute and its corresponding attribute values on FashionAI, Zappos50k, DARN and DeepFashion.}
% Existing works marked with stars (*) are our implementation.}
\label{tab:dataset}
\centering 
\scalebox{1.0}{
\begin{tabular}{c|c|l|c}
\toprule
%Dataset & Attribute & Values\\
\multicolumn{1}{c}{\textbf{Dataset}}&\multicolumn{1}{c}{\textbf{Attribute}}&\multicolumn{1}{c}{\textbf{Values}}&\multicolumn{1}{c}{\textbf{Total}}\\
\hline
\multirow{8}*{\textbf{FashionAI}}
&skirt length & short length, knee length, midi length, ankle length, floor length, invisible & 6\\
\cline{2-4}
&sleeve length & sleeveless, cup sleeves, short sleeves...& 9\\
\cline{2-4}
&coat length & high waist length, regular length, long length...& 8\\
\cline{2-4}
&pant length & short pant, mid length, 3/4 length, cropped pant, full length, invisible & 6\\
\cline{2-4}
&collar design & shirt collar, peter pan, puritan collar, rib collar, invisible & 5\\
\cline{2-4}
&lapel design & notched, collarless, shawl collar, plus size shawl, invisible & 5\\
\cline{2-4}
&neckline design & strapless neck, deep V neckline, straight neck, V neckline, invisible...& 11\\
\cline{2-4}
&neck design & invisible, turtle neck, ruffle semi-high collar, low turtle neck, draped collar & 5\\
\hline
\multirow{4}*{\textbf{Zappos50k}}
&category & shoes, boots, sandals, slippers & 4\\
\cline{2-4}
&gender & women, men, girls, boys & 4\\
\cline{2-4}
&heel height & 1-4in, 5in\&over, flat, under 1in & 7\\
\cline{2-4}
&closure & buckle, pull on, slip on, hook and loop... & 19\\
\hline
\multirow{9}*{\textbf{DARN}}
&clothes category & formal skirt, cotton clothes, lace shirt, small suit, shirt, knitwear, fur\_clothes... & 20\\
\cline{2-4}
&clothes button & pullover, zipper, single breasted type1, single breasted type2... & 13\\
\cline{2-4}
&clothes color & yellow, apricot, flower colors, red, green, white, rose\_red... & 55\\
\cline{2-4}
&clothes length & mid-long, long, short, normal, ultra-long, ultra-short & 7\\
\cline{2-4}
&clothes pattern & solid color, lattice, flower, animal, abstract, floral... & 28\\
\cline{2-4}
&clothes shaoe & slim, straight, cloak, loose, high\_waist, shape1, A-shape... & 11\\
\cline{2-4}
&collat shaoe & polo, shape1, round, shape2, V\_shape, boat\_neck, ruffle\_collar... & 26\\
\cline{2-4}
&sleeve length & long sleeve, three-quarter sleeve, sleeve1, sleeveless... & 8\\
\cline{2-4}
&sleeve shape & puff sleeve, regular, lantern sleeve, pile sleeve... & 17\\
\hline
\multirow{5}*{\textbf{DeepFashion}}
&texture-related & abstract, animal, bandana, baroque, bird, breton, butterfly... & 156\\
\cline{2-4}
&fabric-related & acid, applique, bead, bejeweled, cable, canvas, chenille, chino... & 218\\
\cline{2-4}
&shape-related & a-line, ankle, asymmetric, baja, bermuda, bodycon, box... & 180\\
\cline{2-4}
&part-related & arrow collar, back bow, batwing, bell, button, cinched, collared... & 216\\
\cline{2-4}
&style-related & americana, art, athletic, barbie, beach, bella, blah, boho... & 230\\
\bottomrule
\end{tabular}
 }
\end{table*}
\linespread{1}

%=====================================================================================
\subsection{Compared Models}
%=====================================================================================
The conceptual structures of the five models are demonstrated in Fig. \ref{fig:models}. All of them are based on the same CNN backbone. 

- \textit{Triplet network}: 
This model learns a general embedding space to measure the fine-grained fashion similarity. It simply ignores attributes and performs mean pooling on feature map generated by CNN. The standard triplet ranking loss is used to train the model.

- \textit{Conditional similarity network (CSN)} \cite{veit2017}:
This model first learns an overall embedding space and then employs a fixed mask to select relevant embedding dimensions w.r.t. the specified attribute.

- \textit{ASEN w/o ASA}: 
The model employs mean pooling instead of attribute-aware spatial attention module to aggregate features.

- \textit{ASEN  w/o ACA}: 
The model utilizes the vector $I_s$ as the attribute-specific feature vector without employing the attribute-aware channel attention.

-\textit{ASEN}: It is our proposed model for fine-grained fashion similarity learning.

%=====================================================================================
\subsection{Implementation Details}
%=====================================================================================
On the FashionAI, DARN and DeepFashion dataset, we adopt the same setting. We use ResNet-50 network pre-trained on ImageNet \cite{deng2009imagenet} as the backbone network. Input images are first resized to 224 by 224. The dimension of the attribute-specific embedding is set to be 1024.

On Zappos50K, we follow the same experimental setting as \cite{veit2017} for a fair comparison. Concretely, we utilize ResNet-18 network pre-trained on ImageNet as the backbone network, resize input images to 112 by 112 and set the dimension of the attribute-specific embedding to 64.

On all datasets, the model is trained by ADAM optimizer \cite{kingma2014adam} with an initial learning rate of 1E-4. The learning rate is decayed by multiplying it with 0.985 after each epoch. We train every model for 200 epochs, and the snapshot with the highest performance on validation set is reserved for evaluation on the test set.

% We will release our source code.

\section{More Experiment Results}

\subsubsection{Attribute-Specific Retrieval on DeepFashion}
Table \ref{tab:DeepFashion} summarizes the results on DeepFashion. Our proposed network outperforms the other counterparts. The result again verifies the effectiveness of our proposed ASEN. But the MAP scores on DeepFashion of all models are relatively worse compared to that on FashonAI and DARN. We attribute it to the relative low annotation quality of DeepFashion. For instance, only 77.8\% of images annotated with \textit{A-line} of Shape type are correctly labelled \cite{zou2019fashionai}.

\subsubsection{The Qualitative Results}
\begin{figure*}[tb!]
\centering\includegraphics[width=2\columnwidth]{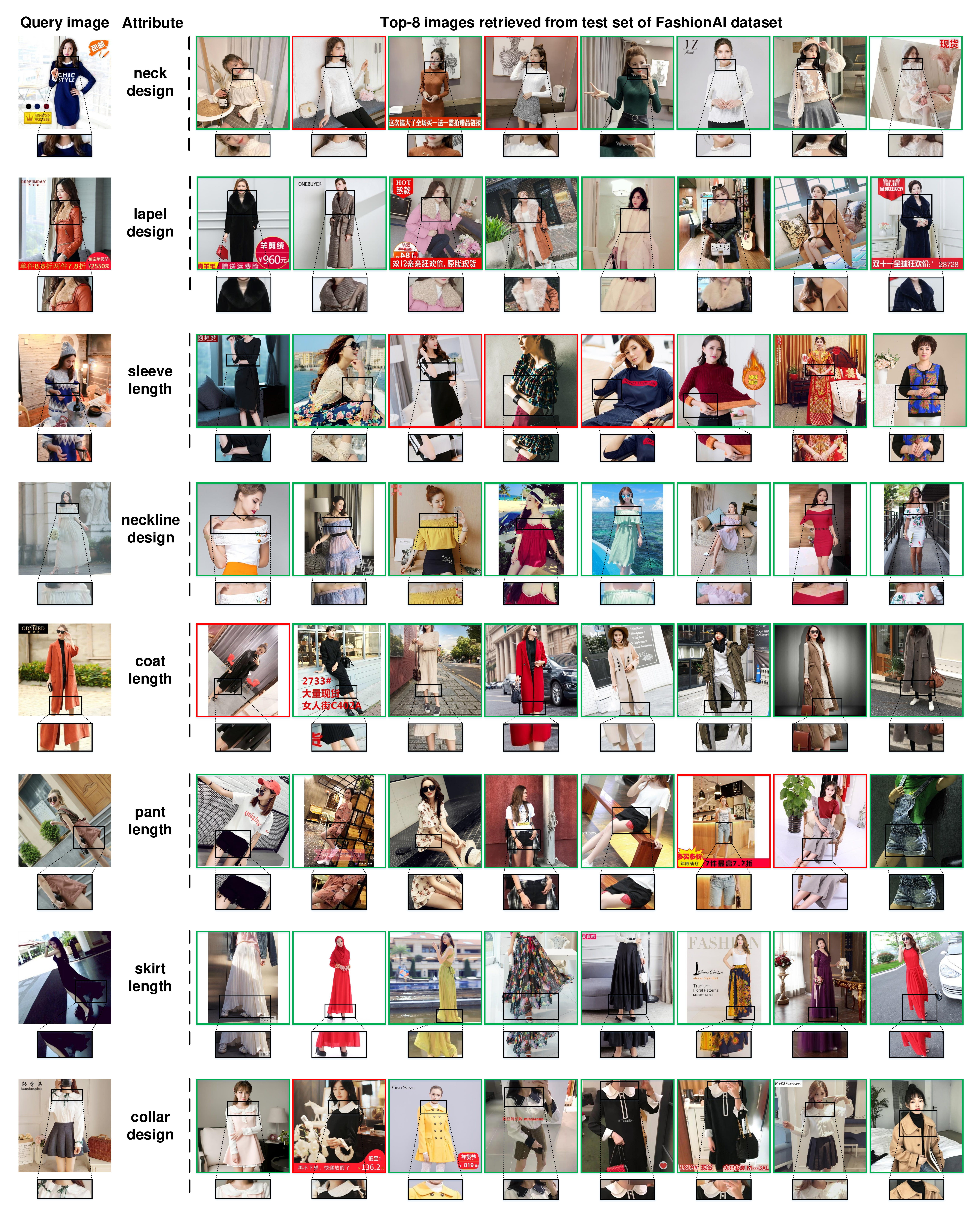}
\caption{Attribute-specific fashion retrieval examples on FashionAI.
Green bounding box indicates the image has the same attribute value with the given image in terms of the given attribute, while the red one indicates the different attribute values.}\label{fig:retrieval_examples_sup}
\end{figure*}

\begin{figure*}[tb!]
\centering\includegraphics[width=2\columnwidth]{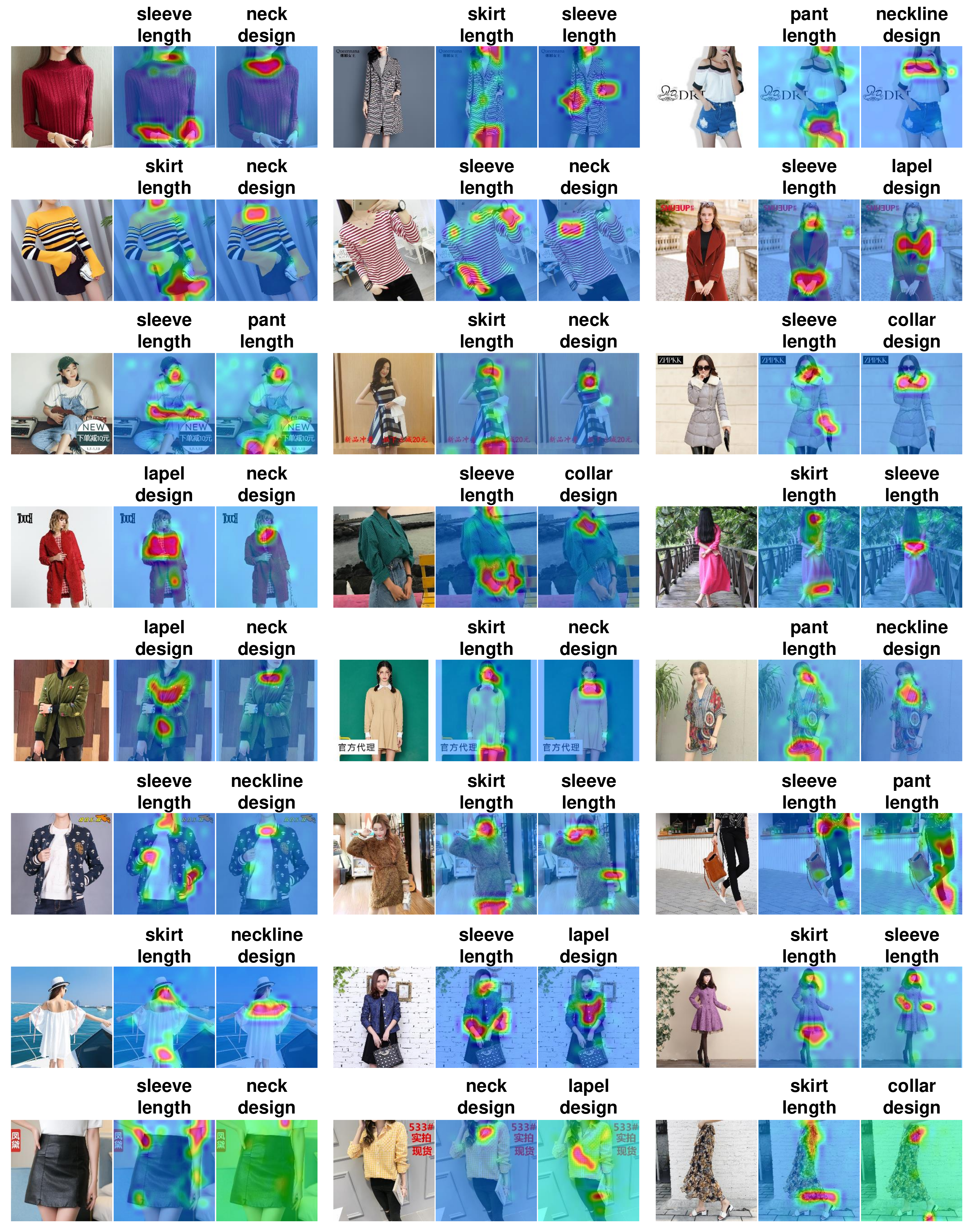}
\caption{Visualization of attribute-specific spatial attention with the guidance of a specified attribute (above the attention image) on FashionAI. 
}\label{fig:spatial_attention_sup}
\end{figure*}

\begin{figure*}[tb!]
\centering\includegraphics[width=2.1\columnwidth]{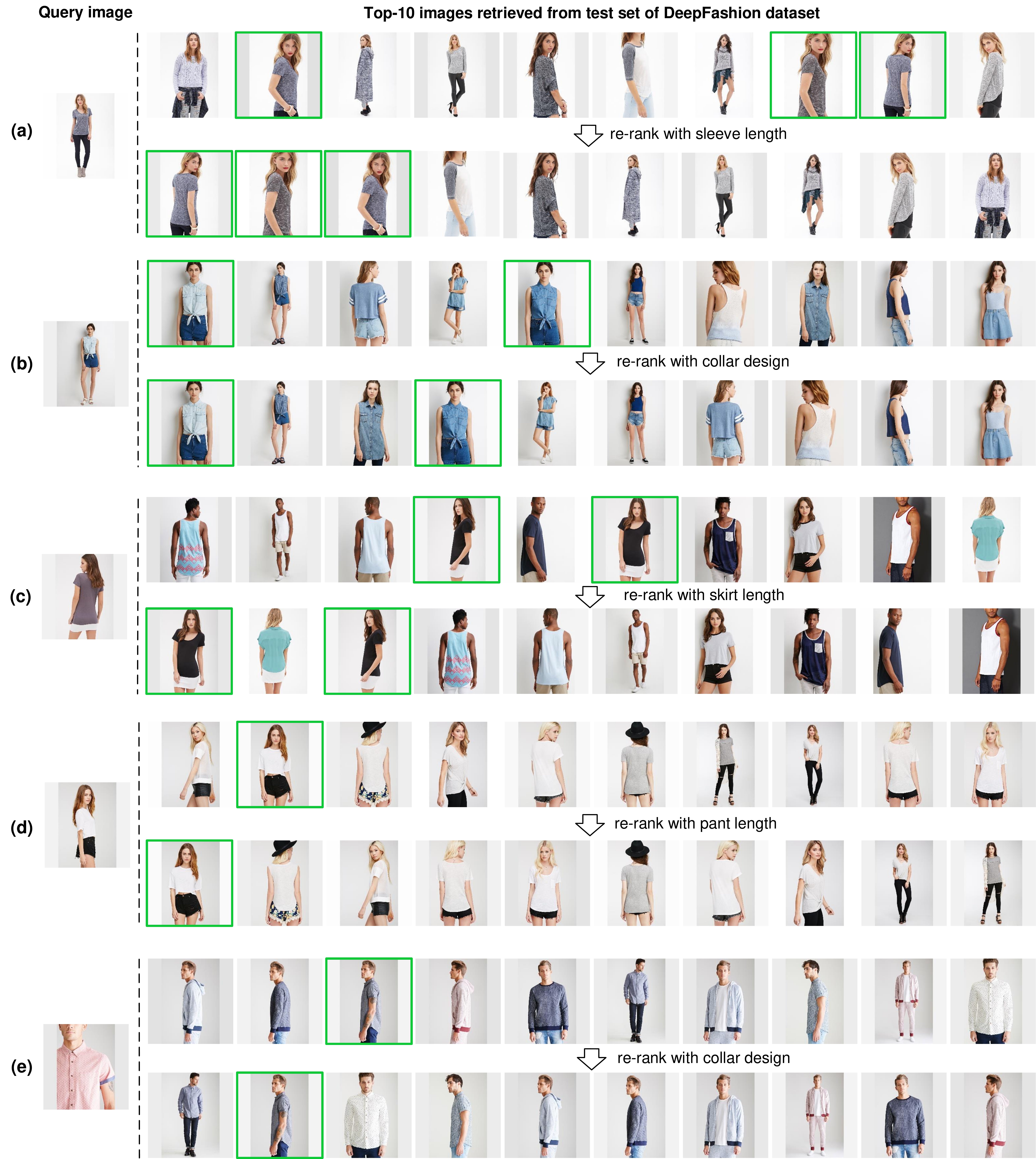}
\caption{Reranking examples for in-shop clothes retrieval on DeepFashion. After the fashion reranking by our proposed ASEN, the results look better.
}\label{fig:re-ranking}
\end{figure*}

\begin{figure*}[tb!]
\centering\includegraphics[width=2.1\columnwidth]{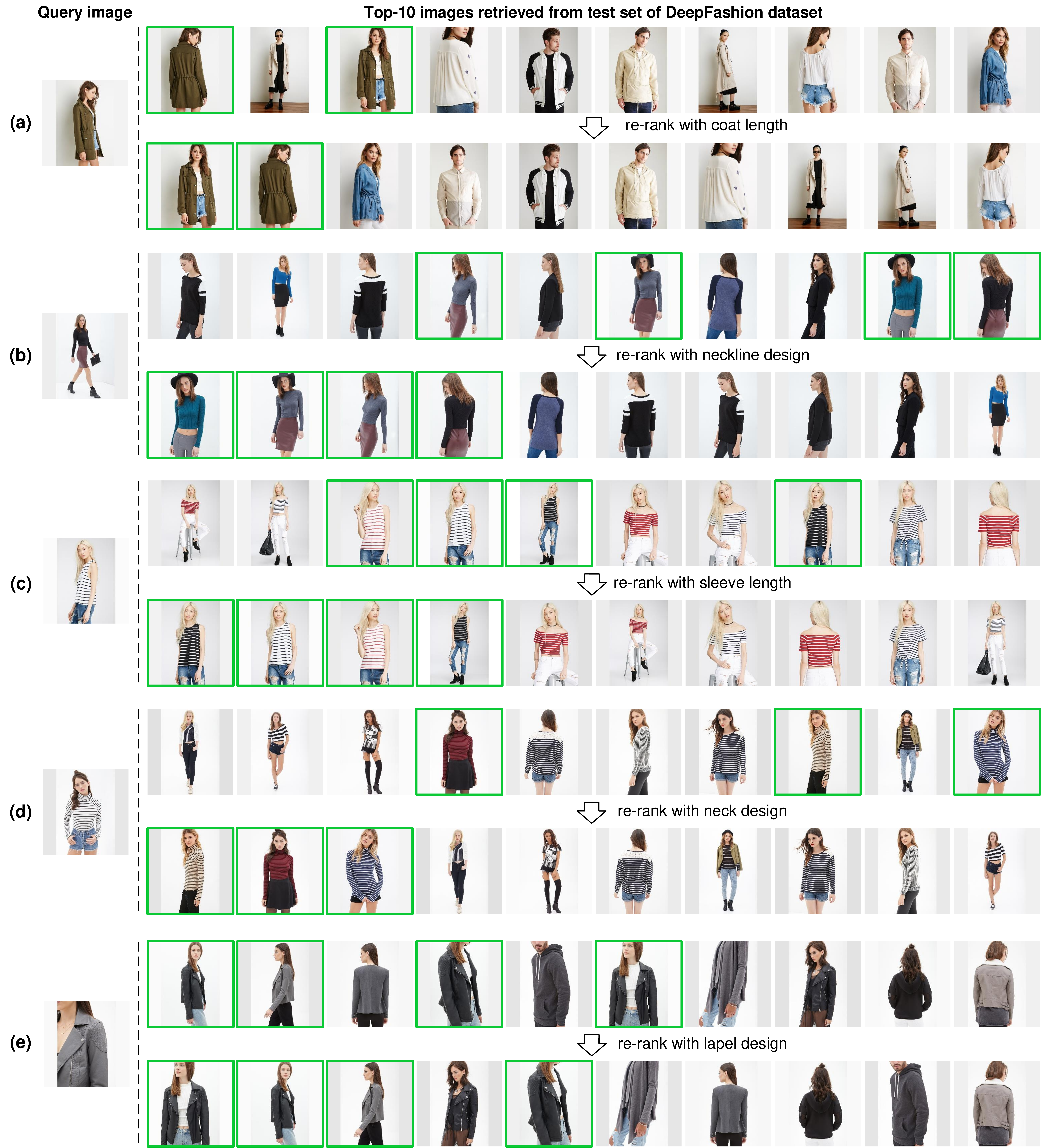}
\caption{Reranking examples for in-shop clothes retrieval on DeepFashion. After the fashion reranking by our proposed ASEN, the results look better.
}\label{fig:re-ranking2}
\end{figure*}

We demonstrate more attribute-specific fashion retrieval examples in Fig. \ref{fig:retrieval_examples_sup}, attention visualization examples in Fig. \ref{fig:spatial_attention_sup} and fashion reranking examples in Fig. \ref{fig:re-ranking} and Fig. \ref{fig:re-ranking2}.

\subsubsection{Efficiency Evaluation}
% Our system runs on a server with the Intel Xeon E5-2680 v4 CPU(@ 2.40GHz) with 14 cores and 252 GB RAM memory, with GeForce GTX 1080 Ti GPUs. It costs about 3 seconds to extract features for 1000 images.
Once our model is trained, given an image and a specific attribute, it takes approximately 0.3 milliseconds to extract the corresponding attribute-aware feature. The speed is fast enough for attribute-based fashion applications.
The performance is tested on a server with an Intel Xeon E5-2680 v4 CPU, 256 G RAM and a GeForce GTX 1080 Ti GPU.

\end{document}